\documentclass[10pt,twocolumn,letterpaper]{article}

\usepackage[cvpr]{cvpr}

\definecolor{cvprblue}{rgb}{0.21,0.49,0.74}
\usepackage[pagebackref,breaklinks,colorlinks,allcolors=cvprblue]{hyperref}

\usepackage{caption}
\usepackage{float}
\usepackage{multirow}

\def\paperID{7811} 
\def\confName{CVPR}
\def\confYear{2025}

\title{PhyT2V: LLM-Guided Iterative Self-Refinement for Physics-Grounded Text-to-Video Generation}

\author{Qiyao Xue, Xiangyu Yin, Boyuan Yang and Wei Gao\\
	University of Pittsburgh\\
	{\tt\small \{qix63, eric.yin, by.yang, weigao\}@pitt.edu}\\
}

\begin{document}
	
	\makeatletter
	\g@addto@macro\@maketitle{
		\begin{figure}[H]
			\setlength{\linewidth}{\textwidth}
			\setlength{\hsize}{\textwidth}
			\centering
			\vspace{-0.3in}
			\includegraphics[width=1\textwidth]{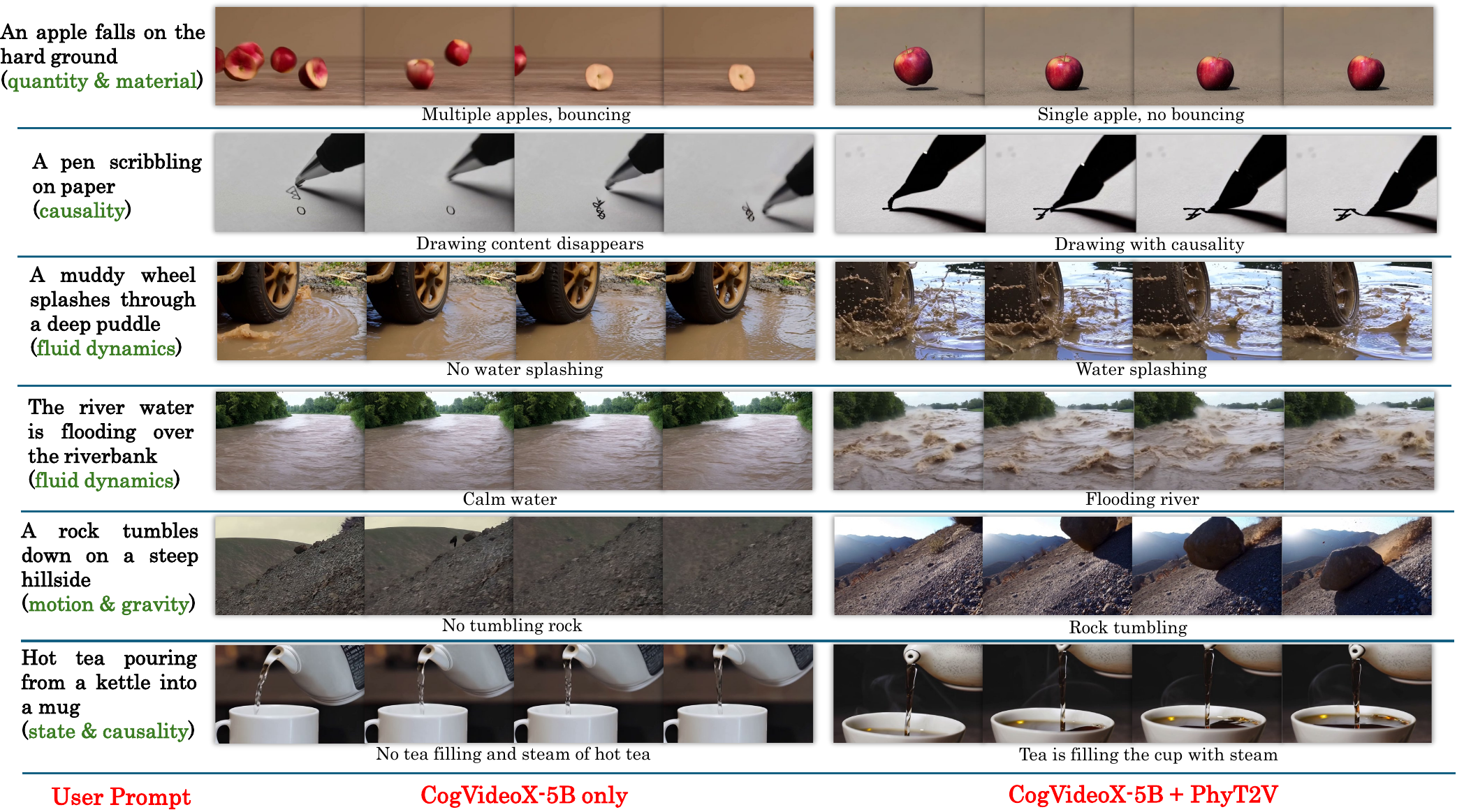} 
			\vspace{-0.2in}
			\caption{\textit{Left}: videos generated by the current text-to-video generation model (CogVideoX-5B \cite{yang2024cogvideoxtexttovideodiffusionmodels}) cannot adhere to the real-world physical rules (described in brackets following the user prompt). \textit{Right}: our method PhyT2V, when applied to the same model, better reflects the real-world physical knowledge.}
			\label{fig:coverdemo}
		\end{figure}
	}
	\makeatother
	
	\maketitle

	\begin{abstract}
		Text-to-video (T2V) generation has been recently enabled by transformer-based diffusion models, but current T2V models lack capabilities in adhering to the real-world common knowledge and physical rules, due to their limited understanding of physical realism and deficiency in temporal modeling. Existing solutions are either data-driven or require extra model inputs, but cannot be generalizable to out-of-distribution domains. In this paper, we present PhyT2V, a new data-independent T2V technique that expands the current T2V model's capability of video generation to out-of-distribution domains, by enabling chain-of-thought and step-back reasoning in T2V prompting. Our experiments show that PhyT2V improves existing T2V models' adherence to real-world physical rules by 2.3x, and achieves 35\% improvement compared to T2V prompt enhancers.
	\end{abstract}
	
	
	\vspace{-0.2in}
	\section{Introduction}
	\vspace{-0.08in}
	\label{sec:introduction}
	
	Text-to-video (T2V) generation has recently marked a significant breakthrough of generative AI, with the advent of transformer-based diffusion models such as Sora \cite{sorabrooks2024video}, Pika \cite{pikalabs} and CogVideoX \cite{yang2024cogvideox} that can produce videos conditioned on textual prompts. These models demonstrate astonishing capabilities of generating complex and photorealistic scenes, and could even make it difficult for humans to distinguish between real-world and AI-generated videos, in the aspect of individual video frames' quality \cite{thompson2024aivideo,luke2024aivideo}.
	
	
	On the other hand, as shown in Figure \ref{fig:coverdemo} - Left, current T2V  models still have significant drawbacks in adhering to the real-world common knowledge and physical rules, such as quantity, material, fluid dynamics, gravity, motion, collision and causality, and such limitations fundamentally prevent current T2V models from being used for real-world simulation \cite{dounas2009blender, qiu2016unrealcv, Kr_ger_2024}. Enforcement of real-world knowledge and physical rules in T2V generation, however, is challenging because it requires the models' understandings of not only individual objects but also how these objects move and interact with each other. Further, unlike generating static images, T2V generation requires frame-to-frame consistency in object appearance, shape, motion, lighting and other dynamics \cite{hong2022cogvideo}. Current T2V models often lack such temporal modeling, especially over long sequences \cite{li2024surveylongvideogeneration}, and the generated videos often contain flickering, inconsistent motion and object deformations across frames \cite{liu2025physgen}. 
	
	Most of the existing solutions to these challenges are \emph{data-driven}, by using large multimodal T2V datasets that cover different real-world domains to train the diffusion model \cite{yang2023learning, gupta2025photorealistic, wang2024worlddreamergeneralworldmodels,huang2024freezeasguard}. However, these solutions heavily rely on the volume, quality and diversity of datasets \cite{yang2024cogvideox,wang2023lavie}. Since real-world common knowledge and physical rules are not explicitly embedded in the T2V generation process, the quality of video generation would largely drop in out-of-distribution domains that are not covered by the training dataset, and the generalizability of T2V models is limited due to the vast diversity of real-world scenario domains. Alternatively, researchers also use the existing 3D engines (e.g, Blender \cite{blendergameengine}, Unity3D \cite{unity3d} and Unreal \cite{unrealengine}) or mathematical models of edge and depth maps \cite{lv2024gpt4motion, lv2024gpt4motionscriptingphysicalmotions, liu2025physgen} to inject real-world physical knowledge into the T2V model, but these approaches are limited to fixed physical categories and patterns such as predefined objects and movements
	\cite{yang2023learning, liu2025physgen}, similarly lacking generalizability.
	
	To achieve generalizable enforcement of physics-grounded T2V generation, we propose a fundamentally different approach: instead of expanding the training dataset or further complicating the T2V model architecture, we aim to expand the current T2V model's capability of video generation from in-distribution to out-of-distribution domains, by embedding real-world knowledge and physical rules into the text prompts with sufficient and appropriate contexts. To avoid ambiguous and unexplainable prompt engineering \cite{gu2023systematicsurveypromptengineering, schulhoff2024promptreportsystematicsurvey, sahoo2024systematicsurveypromptengineering}, our basic idea is to enable chain-of-thought (CoT) and step-back reasoning in T2V generation prompting, to ensure that T2V models follow correct physical dynamics and inter-frame consistency by applying step-by-step guidance and iterative refinement.
	
	\begin{figure}
		\centering
		\includegraphics[width=\linewidth]{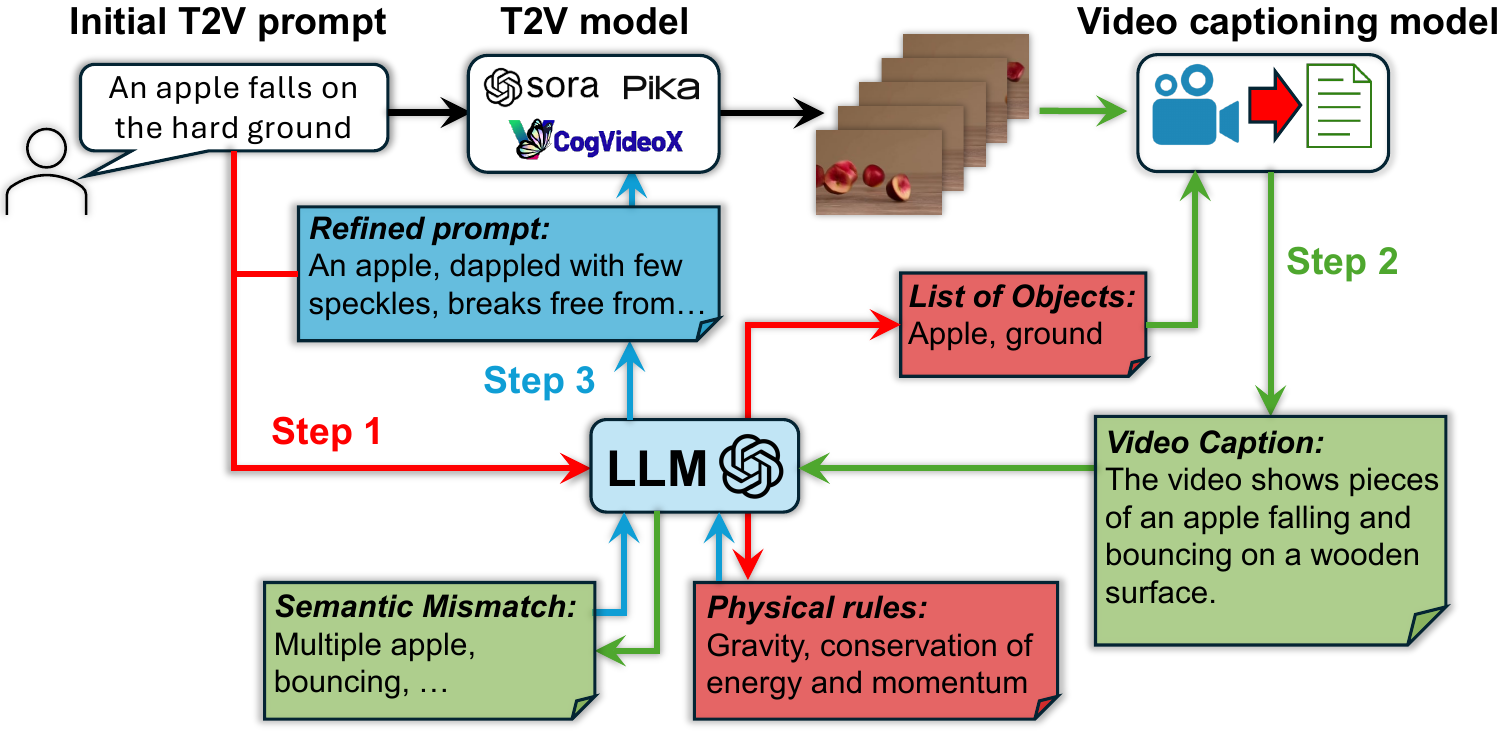} 
		\vspace{-0.2in}
		\caption{One iteration of video and prompt self-refinement in PhyT2V. Such self-refinement will be iteratively conducted in multiple rounds until the quality of generated video is satisfactory.}
		\vspace{-0.2in}
		\label{fig:overview}
	\end{figure}
	
	Based on this idea, this paper presents \textbf{Phy}scial-grounded \textbf{T}ext-\textbf{to}-\textbf{V}ideo (\textbf{PhyT2V}), a new T2V technique that harnesses the natural language reasoning capabilities of well-trained LLMs (e.g, ChatGPT-4o), to facilitate CoT and step-back reasoning as described above. As shown in Figure \ref{fig:overview}, such reasoning is iteratively conducted in PhyT2V, and each iteration autonomously refines both the T2V prompt and generated video in three steps. In Step 1, the LLM analyzes the T2V prompt to extract objects to be shown and physical rules to follow in the video via in-context learning. In Step 2, we first use a video captioning model to translate the video's semantic contents into texts according to the list of objects obtained in Step 1, and then use the LLM to evaluate the mismatch between the video caption and current T2V prompt via CoT reasoning. In Step 3, the LLM refines the current T2V prompt, by incorporating the physical rules summarized in Step 1 and resolving the mismatch derived in Step 2, through step-back prompting. The refined T2V prompt is then used by the T2V model again for video generation, starting a new round of refinement. Such iterative refinement stops when the quality of generated video is satisfactory or the improvement of video quality converges.

	For physcial-grounded video generation performance, we further evaluated PhyT2V by applying it onto multiple SOTA T2V models, by using ChatGPT4 o1-preview \cite{openai2024o1} for LLM reasoning and Tarsier \cite{wang2024tarsier} as the video captioning model. We used two major T2V prompt datasets that cover 7 different real-world domains, and compared PhyT2V with the most competitive baselines of prompt enhancers. Our main findings are as follows:
	
	\begin{itemize}
		\item PhyT2V is highly effective. Without involving any model retraining efforts on any auxiliary model inputs, PhyT2V can improve the adherence of the existing T2V models' generated videos to real-world physical rules by up to 2.3x, by only refining the text prompts to the T2V model.
		\item PhyT2V is high generic. It can result in significant improvement of video quality in a large diversity of real-world domains, covering solid, liquid, mechanics, optics, thermal, etc. It is fully data independent, and its prompting templates can be applied to any existing T2V models with different architectures and input formats.
		\item Based on LLM-guided reasoning and self-refinement, PhyT2V is fully automated and involve the minimum amount of engineering and manual efforts.    
	\end{itemize}
	
	
	\section{Related Work and Motivation}
	\label{sec:related_work}
	\subsection{T2V Generation Models}
	
	Early T2V techniques generate video frames from text-to-image model outputs with temporal extensions \cite{singer2022make}, but cannot maintain temporal consistency and coherence over time, often producing visually appealing but temporally disconnected outputs. Diffusion Transformers (DiT) \cite{peebles2023scalable} improved such consistency with a transformer backbone capable of capturing more complex temporal dynamics and relationships across frames through attention mechanism and long-range dependency modeling \cite{yang2024cogvideox, wang2023lavie}. Based on the DiT architecture, recent T2V models, such as OpenSora \cite{opensora} and VideoCrafter \cite{chen2024videocrafter2}, demonstrated that T2V generation can be further improved by in-context learning when provided with sufficient contextual information \cite{wang2023context}.
	
	\begin{figure}[ht]
		\centering
		\vspace{-0.1in}        
		\includegraphics[width=\linewidth]{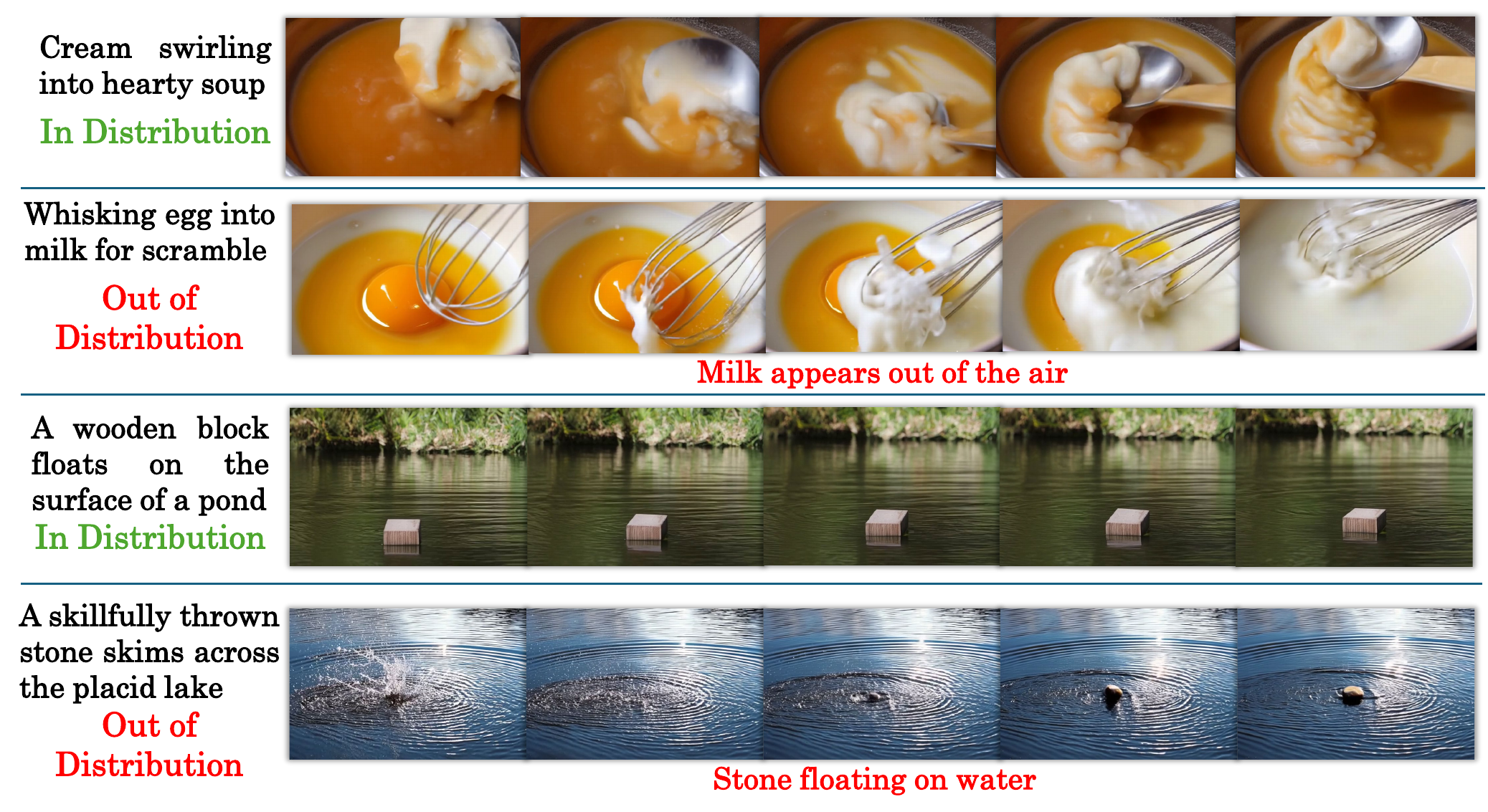} 
		\vspace{-0.25in}
		\caption{Examples of videos generated from in-distribution and out-of-distribution prompts, using the CogVideoX-5B model}
		\vspace{-0.1in}
		\label{fig:example}
	\end{figure}
	
	However, as shown in Figure \ref{fig:example}, although these T2V models demonstrate strong capabilities in video generation when dealing with prompts aligned with the distributions found in the training data, they encounter significant challenges with out-of-distribution prompts that are not covered by training data\footnote{In Figure \ref{fig:example}, the in-distribution prompts are picked from the ones listed in \cite{yang2024cogvideoxtexttovideodiffusionmodels}, and the out-of-distribution prompts are our crafted ones for similar scenarios as the in-distribution prompts.}. In such cases, the outputs often contain physical illusions or artifacts, reflecting the model's limitations in generating realistic and coherent video contents under unfamiliar conditions. Such limitations can be addressed by enlarging the training datasets, improving T2V model architectures or developing new mechanisms for adaptation and error correction \cite{wang2024customvideo,wang2023internvid}, but these approaches are all prompt-specific and lack generalizability.
	
	\begin{figure}[ht]
		\centering
		\includegraphics[width=\linewidth]{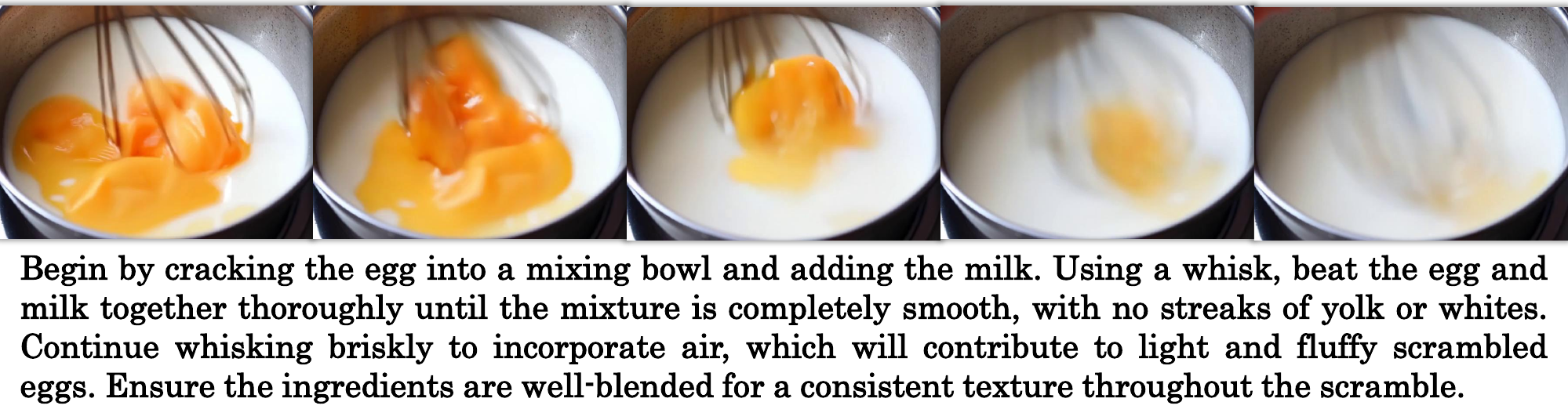} 
		\vspace{-0.25in}
		\caption{A video generated by enhancing the out-of-distribution prompt ``Whisking egg into milk for scramble'' in Figure \ref{fig:example}}
		\vspace{-0.1in}
		\label{fig:goodexample}
	\end{figure}
	
	On the other hand, as shown in Figure \ref{fig:goodexample}, recent research has demonstrated that the quality of video generation with an out-of-distribution prompt can be improved by refining the prompt with sufficient and appropriate details \cite{yang2024cogvideox, hong2022cogvideo}. 
	These findings motivate our design of PhyT2V that embeds contexts of real-world knowledge and physical rules into T2V prompts, to guide the T2V process for better physical accuracy and temporal alignment. The existing works, however, could still fail when tackling more intricate scenarios such as multi-object interactions, because the T2V model lacks an efficient feedback mechanism to learn how the generated video deviates from the real-world knowledge and physical rules. Researchers suggest to provide such feedback with extra input modalities to T2V models such as sampled video frames, depth map or scribble images \cite{wang2023context, zhang2024video}, but incur significant amounts of extra computing overhead and cannot be generalizable. Instead, in our design of PhyT2V, we aim to fully automate the feedback with only text prompts, and enable iterative feedback for the optimum video quality.

	\begin{figure}[ht]
		\centering
		\vspace{-0.1in}        
		\includegraphics[width=0.95\linewidth]{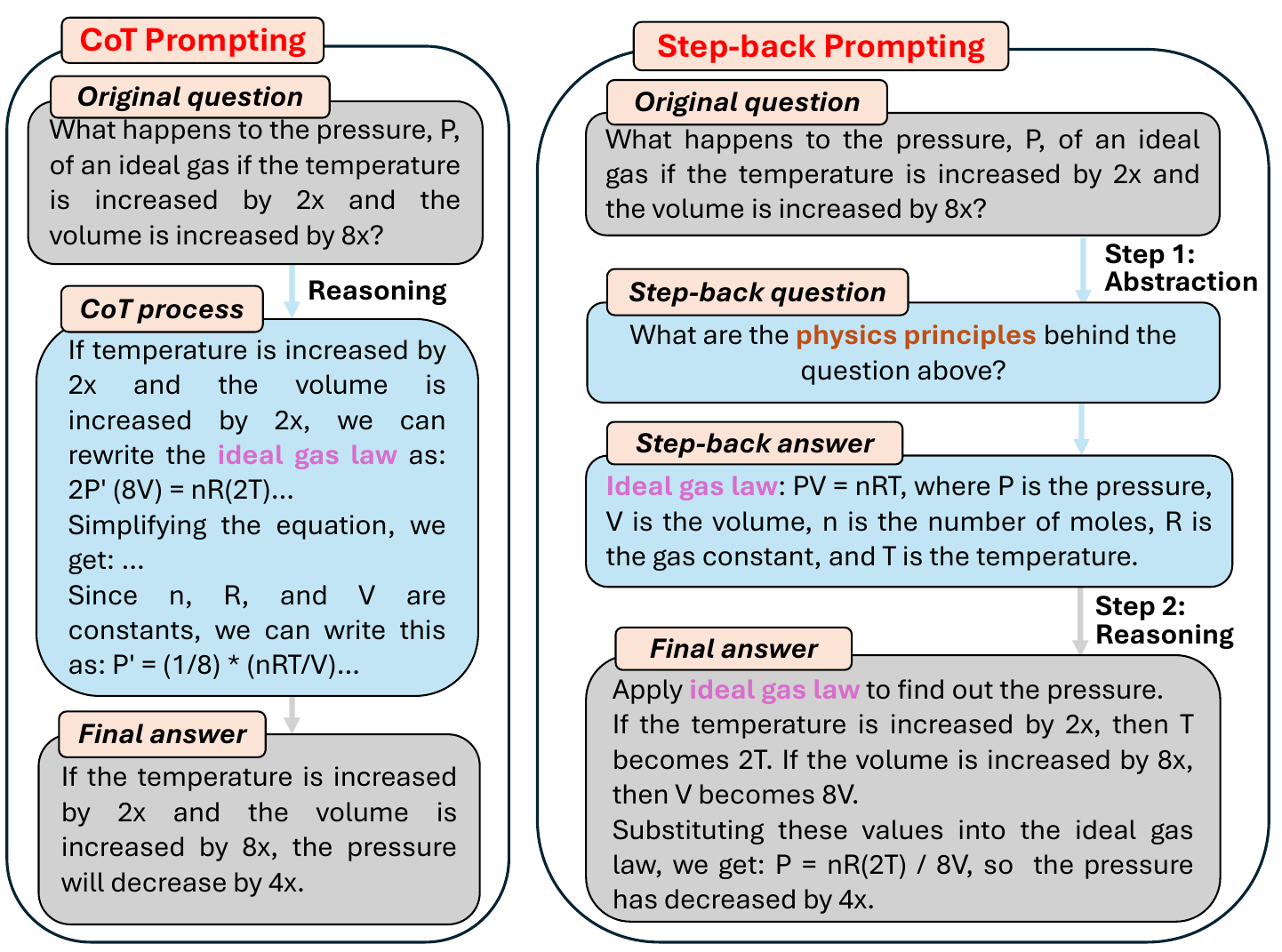} 
		\vspace{-0.1in}
		\caption{Examples of CoT and step-back reasoning}
		\vspace{-0.1in}
		\label{fig:CoT&Step-back_example}
	\end{figure}
	
	\begin{figure*}[ht]
		\centering
		\includegraphics[width=0.96\textwidth]{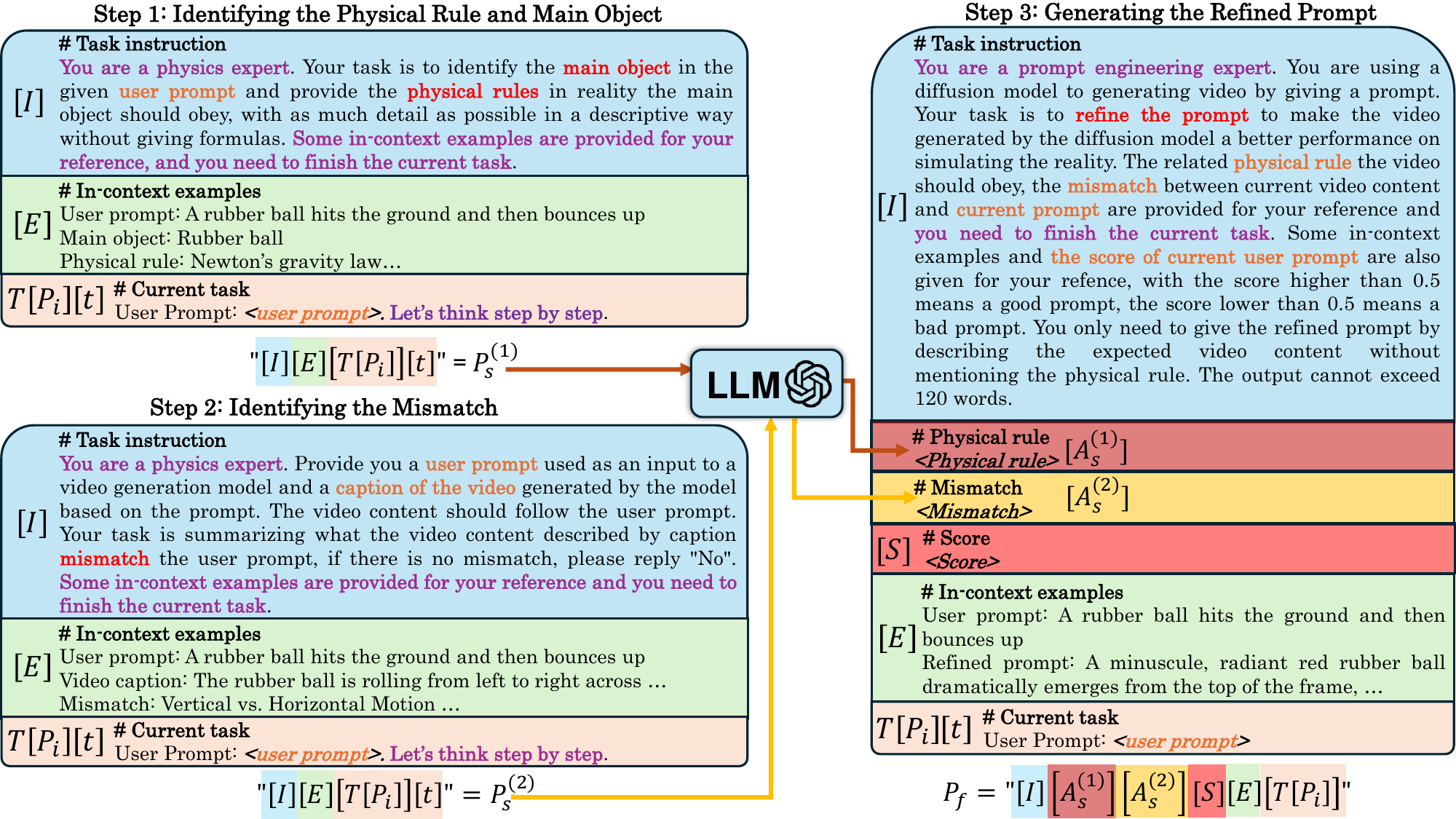} 
		\vspace{-0.05in}
		\caption{Our design of PhyT2V, illustrated by one round of video refinement consisting of three steps. Texts in brown are inputs from previous step. Texts in red are outputs to the next step; Texts in purple are prompts to trigger LLM reasoning}
		\vspace{-0.2in}
		\label{fig:prompt}
	\end{figure*}
	
	 \vspace{-0.05in}
	\subsection{Using LLM in T2V Generation}
	\vspace{-0.05in}
	
	LLMs with strong capabilities in natural language processing (NLP) have been a natural choice for prompt refinement in text-to-image and text-to-video generation, and existing work utilized LLMs to interpret text prompts and orchestrate the initial layout configurations \cite{lian2023llmimage, lian2023llmvideo, lin2023videodirectorgpt, zhu2024compositional,wu2024self, yang2024mastering,huang2024towards,huang2025mpnp}. However, since current LLMs lack inherent understandings of the real-world physical laws, using LLMs with simple instructions usually result in videos that appear visually coherent but lack accurate physical realism, particularly when generating scenes with complex object interactions. Furthermore, these approaches frequently rely on static prompts or simple iterative refinements based on bounding box and segmentation map, which may capture basic visual attributes but fail to adapt to nuanced changes that require continuous physical modeling and adjustment.
	
	An effective approach to addressing these limitations and providing effective feedback for prompt refinement is to explicitly trigger in-context learning and reasoning in LLM. For example, as shown in Figure \ref{fig:CoT&Step-back_example}, CoT reasoning deconstructs complex prompts into stepwise logical tasks, and hence ensures a precise scheduling path to align generated content with the input prompt. However, CoT reasoning, in some cases, could make errors in some intermediate steps, and step-back prompting can address this limitation by further deriving the step-back question at a higher level of abstraction and hence avoiding confusions and vagueness. In our design of PhyT2V, we will utilize such LLM reasoning to analyze the inconsistency of the generated video to real-world common knowledge and physical rules, and use the reasoning outcome as feedback for T2V prompt refinement.
	
	The Chain-of-Thought (CoT) method is suitable for single data mode processing because it emphasizes linear decomposition and step-by-step reasoning, and it is especially effective for data processing flows that do not require cross-modal synchronization or interaction. However, multimodal data processing involves the fusion of data from different modalities and complex synchronization requirements, which cannot be completed through simple linear decomposition and requires frequent cross-modal information interaction and parallel processing, which exceeds the linear reasoning ability of the CoT method, resulting in limited performance in multimodal tasks. This is the reason why it is hard  to directly apply CoT reasoning in T2V process itself as it requires multimodal alignment between the next and video modality. This also motivates us to adopt video captioning and use the video caption in the reasoning process, so that we can conduct CoT and step-back reasoning only in the text domain.
	
	\vspace{-0.05in}
	\section{Method}
	\vspace{-0.05in}
	\label{sec:method}
	In this section, we present details of our PhyT2V design. 
	In principal, PhyT2V's refinement of T2V generation is an iterative process consisting of multiple rounds. In each round, as shown in Figure \ref{fig:prompt}, the primary objective of our PhyT2V design is to guide a well-trained LLM (e.g., ChatGPT-4o) to generate a refined prompt that enables the pre-trained T2V model to generate videos that better match the given user prompt and real-world physical rules, and the refined prompt will be iteratively used as the new user prompt in the next round of refinement. 
	
	Each round of refinement is structured  around decomposing the complex refinement problem into a series of simpler subproblems, more specifically, two parallel subproblems and one final subproblem. The two parallel subproblems are: \emph{Step 1)} identifying the relevant physical rules that the generated video should follow based on the user prompt, and \emph{Step 2)} identifying semantic mismatches between the user prompt and the generated video. Based on the knowledge about physical rules and semantic mismatches, the final subproblem (Step 3) generates the refined prompt to better adhere to the physical rules and resolve the mismatches. 
	
	To ensure proper identification in the parallel subproblems and prompt generation in the final subproblem, the core of PhyT2V design is two types of LLM reasoning processes within the prompt enhancement loop: the \emph{local CoT reasoning} for individual subproblems and \emph{global step-back reasoning} for the overall prompt refinement problem.
	
	\noindent\textbf{Local CoT reasoning} is executed within the prompt for each subproblem, to prompt the LLM to generate a detailed reasoning chain in its latent embedding space \cite{wang2022towards}. 
	Addressing the parallel subproblems facilitates LLM with a more concentrated attention on prerequisites of prompt refinement, enabling a deeper comprehension of the physical laws that govern the video content as well as the identification of discrepancies between the generated video and the user prompt. The outcomes derived from these parallel subproblems reflect the language model's abstraction in step-back reasoning on the overarching prompt refinement.
	
	\noindent\textbf{Global step-back reasoning}: To integrate various subproblems into a coherent framework for prompt and video refinement, one intuitive approach involves employing CoT reasoning across these subproblems, allowing the LLM to engage in self-questioning. However, this method may lead to the risk of traversing incorrect reasoning pathways. Instead, we apply global step-back reasoning across subproblems, by using a self-augmented prompt to incorporate the LLM-generated responses to high-level questions about physical rules and semantic mismatches in earlier parallel problems, when generating the refined prompt in the final subproblem. In this way, we can improve the correctness of intermediate reasoning steps in CoT reasoning, and enable consistent improvement across steps in reasoning.
	
	Both reasoning processes are facilitated through appropriate task instruction prompting tailored to different subproblems. In general, our prompting procedure follows the prompt modeling in \cite{schulhoff2024prompt}, which divides task instructions into several components. More details about these components in our design of PhyT2V are elaborated as follows.
	
	Compared to the previous prompt enhancing methods, PhyT2V's key contribution is to analyze the semantic mismatch between currently 
	generated video and the prompt, as well as refinements based on such mismatch. Previous methods can be formulated as $p' = f_{enhance}(p, \theta)$, where $p$ and $p'$ are the original and enhanced prompts, $f_{\text{enhance}}$ is the prompt enhancer, and $\theta$ represents parameters or rules guiding the enhancement. In contrast, PhyT2V further involves the additional information about the T2V process, i.e., $p' = f_{\text{enhance}}(p, f_{mismatch}(C(V(p)), p), f_{phy}(p), \theta)$, where $f_{phy}(p)$ analyzes the physical rules to be followed given $p$, $V(p)$ is the currently generated video given prompt $p$, $C$ is the video captioning model and $f_{mismatch}$ finds the semantic mismatch between $C(V(p))$ and $p$. The key advantages are: (1) Semantic awareness: the refinement process explicitly incorporates the semantic mismatch to enable targeted T2V improvements; (2) Physical-world knowledge integration: physical rules derived from $p$ enable guided enhancement; (3) Guided reasoning: unlike prior methods that rely solely on templates or embeddings, PhyT2V dynamically adapts prompt refinement to the semantic mismatch.  
	
	\subsection{Prompting in Parallel Subproblems for Local CoT Reasoning}
	\vspace{-0.05in}
	
	In both Step 1 and Step 2, the first part of prompt is a task instruction prompt $[I]$ to instruct the LLM to understand the task in the subproblem. $[I]$ is designed with multiple components, each of which corresponds to different functions. In the first sentence, it provides general guidance to relate the current subproblem to the entire refinement problem, to better condition the subproblem answer. Afterwards, it will include detailed descriptions of the task: identifying the physical rule and main object in Step 1, and identifying the semantic mismatch between the user prompt and caption of the generated video (generated by the video captioning model) in Step 2. It will also contain the requirements about the expected information in LLM's output. For example, in Step 1, the LLM's output about the physical rule should be in a descriptive way without giving formulas.
	
	Besides, to ensure proper CoT reasoning, we follow the existing work \cite{wang2023knowledge,li2024dissecting} and provide in-context examples $[E]$ about tasks. To facilitate LLM's in-context learning \cite{dong2022survey,dong2024survey}, $[E]$ is given in the format of QA pairs. That is, instead of fine-tuning a separate LLM checkpoint for each new task, we prompt the LLM with a few input-output exemplars, to demonstrate the task and condition the task's input-output format to the LLM, to guide the LLM's reasoning process.

	Then, the final part of the prompt, denoted as $[T]$, is the information of the current instance of the task, usually with the current user prompt ($P_i$) being embedded. As a common practice of CoT reasoning, it also contains the hand-crafted trigger phrase ($t$), ``Let's think step by step'', to activate the local CoT reasoning in LLM.

	
	\subsection{Prompting in the Final Subproblem for Global Step-Back Reasoning}
	\vspace{-0.05in}
	In the final subproblem, we enforce global step-back reasoning, by using the outputs of the two parallel subproblems above, i.e., knowledge about the physical rules and the prompt-video mismatch, as the high-level concepts and facts. Grounded on such high-level abstractions, we can make sure to improve the LLM's ability in following the correct reasoning path of generating the refined prompt.
	
	Being similar to the prompts used in the two parallel subproblems above, the prompt structure in the final subproblem also contains $[I]$, $[E]$ and $[T]$. Furthermore, to ensure the correct reasoning path, we also provide quantitative feedback to the LLM about the effectiveness of previous round's prompt refinement. Such effectiveness could be measured by the existing T2V evaluators, which judge the semantic alignment and quality of physical common sense of the currently generated video\footnote{This video is generated using the prompt refined in the previous round, and is also used to generate the video caption as the input in Step 2.}. For example, the VideoCon-Physics evaluator \cite{bansal2024videophy} gives a score ($[S]$) between 0 and 1. If $[S]$ is $<$0.5, it indicates that the refined prompt produced in the previous round is ineffective, hence guiding the LLM to take another alternative reasoning path.
	
	Since the prompt in the final subproblem  is rich with reasoning and inherently very long-tailed, we removed the trigger prompt $[t]$, to prevent incorporating the information in the final answer unrelated to the user's initial input prompt.
	
	\subsection{The Stopping Condition}
	\vspace{-0.05in}
	The process of iterative refinement normally continues until the quality of the generated video is satisfactory, measured by the T2V evaluator as described above. Furthermore, the current T2V models naturally have limitations in generating some complicated or subtle scenes. In these cases, it would be difficult, even for PhyT2V, to reach physical realism after multiple rounds of iterations, and PhyT2V's refinement would stop when the iterations converge, i.e., the improvement of video quality becomes little over rounds.
	
	\vspace{-0.05in}
	\section{Experiments}
	\vspace{-0.05in}
	\noindent\textbf{Models \& Datasets:} We applied PhyT2V on several DiT-based open-source T2V models, as listed below, and evaluated how PhyT2V improves the generated videos' adherence to real-world knowledge and physical rules. We use ChatGPT4 o1-preview \cite{openai2024o1} as the LLM for reasoning, and Tarsier \cite{wang2024tarsier} as the video captioning model. All generated videos last 6 seconds with 10 FPS and resolution of 720$\times$480. Details of evaluation setup are in Appendix A.


	\begin{itemize}
		\item\textbf{CogVideoX \cite{yang2024cogvideox}}: It generates 10-second videos from text prompts, with 16 FPS and 768$\times$1360 resolution. It offers two model variants with 2B and 5B parameters.
		\item\textbf{OpenSora 1.2 \cite{opensora}}: As an alternative to OpenAI's Sora \cite{sorabrooks2024video}, it contains 1.1B parameters and produces videos with 16 seconds, 720p resolution and different aspect ratios.
		\item\textbf{VideoCrafter \cite{chen2024videocrafter2}}: With 1.8B parameters, it can generate both images and videos from text prompts, with 576$\times$1024 resolution and a focus on video dynamics.
	\end{itemize}
	
	\noindent Since we target enhancing the T2V models' capability of generating physics-grounded video contents, we use the following two prompt benchmarks that emphasize physical laws and adherence as the text prompts for T2V:
	
	\begin{itemize}
		\item\textbf{VideoPhy \cite{bansal2024videophy}} is designed to assess whether the generated videos follow physical common sense for real-world activities. It consists 688 human-verified captions that describe interactions between various types of real-world objects, including solid and fluid.    
		\item\textbf{PhyGenBench \cite{meng2024towards}}, similarly, allows evaluating the correctness of following physical common sense in T2V generation. It comprises 160 carefully crafted prompts spanning four physical domains, namely mechanics, optics, thermal and material properties. Since the domain of material properties has been covered by VideoPhy, we use the first three domains listed above.    
	\end{itemize}
	
	\noindent\textbf{Evaluation metric:} We use VideoCon-Physics evaluator provided with VideoPhy \cite{bansal2024videophy}, to measure how the generated video adheres to physical common sense (PC) and achieves semantic adherence (SA). The PC metric evaluates whether the depicted actions and object's state follow the real-world physics laws. The SA metric measures if the actions, events, entities and their interactions specified in the prompt are present. Both metrics yield binary outputs: 1 indicates adherence and 0 indicates otherwise. On each T2V model and dataset, the binary outputs from all prompts are averaged. 
	
	In addition, we also evaluated PhyT2V using the widely used VBench metrics and benchmarks \cite{huang2024vbench}, which allow comprehensive evaluations of the generated video in multiple aspects, including video quality, video-condition consistency, prompt following and human preference alignment.
	
	\noindent\textbf{Baselines:} For fair comparison, we only use the existing T2V prompt enhancers as baselines, and other existing work with extra inputs to T2V models \cite{dounas2009blender, qiu2016unrealcv, Kr_ger_2024,lv2024gpt4motion, liu2025physgen}
	are not applicable. We involve two prompt enhancers: 1) Directly using the existing LLM, particularly ChatGPT4, as the prompt enhancer \cite{lv2024gpt4motionscriptingphysicalmotions,yang2024worldgpt}; 2) Promptist \cite{li2024promptist}, which uses reinforcement learning to automatically refine and enhance prompts in the model-preferred way.

	\vspace{-0.05in}
	\subsection{Improvement of the Generated Video Quality}
	\vspace{-0.05in}
	\label{subsec:main_results}
	
	As shown in Table \ref{tab:main_results_videophy} and \ref{tab:main_results_phygenbench}, when PhyT2V is applied to different T2V models, it can significantly improve the generated video's adherence to both the text prompt itself and the real-world physical rules, compared to the videos generated by vanilla T2V models (i.e., in Round 1 of PhyT2V's refinement). In particular, such improvement is the most significant on the CogVideoX-2B model, where PC improvement can be up to 2.2x and SA improvement can be up to 2.3x. On all the other models, PhyT2V can also reach noticeable improvement, ranging from 1.3x to 1.9x.
	
	\begin{table}
		\centering
		\vspace{-0.1in}
		{\fontsize{7.5}{9}\selectfont
			\begin{tabular}{ccccccc}
				\toprule
				\multicolumn{2}{c}{\textbf{Round}} & \textbf{1} & \textbf{2} &\textbf{3} &\textbf{4} & \\
				\midrule[1pt]
				\multirow{2}{*}{\textbf{CogVideoX-5B \cite{yang2024cogvideox}}}
				&\textbf{PC}&0.26&0.32&0.39 & 0.42 &\\ \cmidrule(ll){2-7}
				&\textbf{SA}&0.48&0.52&0.56& 0.59 &\\ 
				\midrule[1pt]
				\multirow{2}{*}{\textbf{CogVideoX-2B \cite{yang2024cogvideox}}}
				&\textbf{PC}&0.13&0.19&0.27&0.29&\\ \cmidrule(ll){2-7}
				&\textbf{SA}&0.22&0.12&0.40&0.42&\\
				\midrule[1pt]
				\multirow{2}{*}{\textbf{OpenSora \cite{opensora}}}
				&\textbf{PC}&0.17& 0.29& 0.27& 0.31\\ \cmidrule(ll){2-7}
				&\textbf{SA}&0.29& 0.38& 0.44& 0.47\\  
				\midrule[1pt]
				\multirow{2}{*}{\textbf{VideoCrafter \cite{chen2024videocrafter2}}}
				&\textbf{PC}&0.15&0.25&0.29&0.33&\\ \cmidrule(ll){2-7}
				&\textbf{SA}&0.24&0.38&0.44&0.49&\\  
				\bottomrule
		\end{tabular}}
		\vspace{-0.1in}
		\caption{The quality of videos generated by different T2V models using the VideoPhy prompt dataset, over multiple rounds of iterative refinement in PhyT2V }
		\label{tab:main_results_videophy}
		\vspace{-0.15in}
	\end{table}
	
	\begin{table}
		\centering
		{\fontsize{7.5}{9}\selectfont
			\begin{tabular}{ccccccc}
				\toprule
				\multicolumn{2}{c}{\textbf{Round}} & \textbf{1} & \textbf{2} &\textbf{3} &\textbf{4} \\
				\midrule[1pt]
				\multirow{2}{*}{\textbf{CogVideoX-5B \cite{yang2024cogvideox}}}
				&\textbf{PC}&0.28&0.32&0.38&0.42&  \\ \cmidrule(ll){2-7}
				&\textbf{SA}&0.22&0.35&0.36&0.38&\\ 
				\midrule[1pt]
				\multirow{2}{*}{\textbf{CogVideoX-2B \cite{yang2024cogvideox}}}
				&\textbf{PC}&0.16&0.19&0.24&0.27&  \\ \cmidrule(ll){2-7}
				&\textbf{SA}&0.15&0.29&0.33&0.35&\\
				\midrule[1pt]
				\multirow{2}{*}{\textbf{OpenSora \cite{opensora}}}
				&\textbf{PC}&0.21&0.25&0.24&0.26&  \\ \cmidrule(ll){2-7}
				&\textbf{SA}&0.23&0.28&0.29&0.30\\  
				\midrule[1pt]
				\multirow{2}{*}{\textbf{VideoCrafter \cite{chen2024videocrafter2}}}
				&\textbf{PC}&0.20&0.24&0.32&0.36  \\ \cmidrule(ll){2-7}
				&\textbf{SA}&0.27&0.33&0.37&0.42 \\  
				\bottomrule
		\end{tabular}}
		\vspace{-0.1in}
		\caption{The quality of videos generated by different T2V models using the PhyGenBench prompt dataset, over multiple rounds of iterative refinement in PhyT2V}
		\vspace{-0.2in}
		\label{tab:main_results_phygenbench}
	\end{table}
	
	Meanwhile, results in Table \ref{tab:main_results_videophy} and \ref{tab:main_results_phygenbench} showed that PhyT2V's process of iterative refinement converge quickly and only takes few rounds. Most improvement of video quality happens in the first two rounds, and little improvement can be observed in the fourth round. Hence, in practice, we believe that 3-4 iterative rounds would be sufficient.
	
	Furthermore, as shown in Table \ref{tab:prompt_enhancer_videophy} and \ref{tab:prompt_enhancer_phygenbench}, PhyT2V also largely outperforms the existing prompt enhancers by at least 35\%, when being applied to CogVideoX-5B and OpenSora models. In particular, ChatGPT 4, when being used as the prompt enhancer, delivers better performance than Promptist due to its stronger language processing capabilities, but still cannot ensure physics-grounded T2V, due to the lack of explicit reasoning on text-to-video alignment.

	\begin{table}
		\centering
		\vspace{-0.05in}
		{\fontsize{7.5}{9}\selectfont
			\begin{tabular}{cccc}
				\toprule		
				\multicolumn{2}{c}{} & \textbf{CogVideoX-5B} & \textbf{OpenSora} \\
				\midrule[1pt]
				\multirow{2}{*}{\textbf{ChatGPT 4 \cite{lv2024gpt4motionscriptingphysicalmotions}}} &\textbf{PC}& 0.33 & 0.21\\
				\cmidrule(ll){2-4}
				&\textbf{SA}& 0.41 & 0.32\\
				\midrule[1pt]
				\multirow{2}{*}{\textbf{Promptist \cite{li2024promptist}}} &\textbf{PC}& 0.25 & 0.19  \\
				\cmidrule(ll){2-4}
				&\textbf{SA}& 0.39 & 0.33 \\
				\bottomrule
		\end{tabular}}
		\vspace{-0.1in}
		\caption{The quality of videos generated by enhancing the prompts in the VideoPhy dataset using different prompt enhancers}
		\label{tab:prompt_enhancer_videophy}
	\end{table}

	\begin{table}
		\centering
		\vspace{-0.1in}
		{\fontsize{7.5}{9}\selectfont
			\begin{tabular}{cccc}
				\toprule		
				\multicolumn{2}{c}{} & \textbf{CogVideoX-5B} & \textbf{OpenSora} \\
				\midrule[1pt]
				\multirow{2}{*}{\textbf{ChatGPT 4 \cite{lv2024gpt4motionscriptingphysicalmotions}}} &\textbf{PC}& 0.27 & 0.20 \\
				\cmidrule(ll){2-4}
				&\textbf{SA}& 0.23 & 0.23\\
				\midrule[1pt]
				\multirow{2}{*}{\textbf{Promptist \cite{li2024promptist}}} &\textbf{PC}& 0.32 & 0.19 \\
				\cmidrule(ll){2-4}
				&\textbf{SA}& 0.24 & 0.21\\
				\bottomrule
		\end{tabular}}
		\vspace{-0.1in}
		\caption{The quality of videos generated by enhancing the prompts in the PhyGenBench dataset using different prompt enhancers}	
		\vspace{-0.2in}
		\label{tab:prompt_enhancer_phygenbench}
	\end{table}

	\begin{table*}[ht]
		\centering
		{\fontsize{7.5}{9}\selectfont
			\begin{tabular}{cc||cccc||cccc||cccc||cccc}
				\toprule
				\multicolumn{2}{c||}{} & \multicolumn{4}{c||}{\textbf{CogVideoX-5B}} & \multicolumn{4}{c||}{\textbf{CogVideoX-2B}} & \multicolumn{4}{c||}{\textbf{OpenSora}} & \multicolumn{4}{c}{\textbf{VideoCrafter}}\\
				\midrule[1pt]
				\multicolumn{2}{c||}{
					\textbf{Round}} & \textbf{1} & \textbf{2} &\textbf{3} & \textbf{4} &\textbf{1} & \textbf{2} &\textbf{3} & \textbf{4} & \textbf{1} & \textbf{2} &\textbf{3} & \textbf{4} & \textbf{1} & \textbf{2} &\textbf{3} & \textbf{4}\\
				\midrule[1pt]
				\multirow{2}{*}{\textbf{Solid-Solid}}
				&\textbf{PC}& 0.21 & 0.28 & 0.34 & 0.32 & 0.09 & 0.13 & 0.14 & 0.22 & 0.12 & 0.27 & 0.29 & 0.30  & 0.19 & 0.22 & 0.27 & 0.28 \\ \cmidrule(ll){2-18}
				&\textbf{SA}& 0.24 & 0.48 & 0.49 & 0.47 & 0.18 & 0.25 & 0.36 & 0.33 & 0.16 & 0.34 & 0.37 & 0.35 & 0.24 & 0.40 & 0.45 & 0.47\\\midrule[1pt]	
				\multirow{2}{*}{\textbf{Solid-Fluid}}
				&\textbf{PC}& 0.22 & 0.27 & 0.28 & 0.30 & 0.11 & 0.18 & 0.28 & 0.27 & 0.17 & 0.21 & 0.24 & 0.25 & 0.18 & 0.24 & 0.25 & 0.26\\ \cmidrule(ll){2-18}
				&\textbf{SA}& 0.39 & 0.54 & 0.60 & 0.61 & 0.29 & 0.43 & 0.44 & 0.43 & 0.16 & 0.40 & 0.41 & 0.36 & 0.34 & 0.43 & 0.48 & 0.52\\  \midrule[1pt]
				\multirow{2}{*}{\textbf{Fluid-Fluid}}
				&\textbf{PC}& 0.57 & 0.59 & 0.63 & 0.62 & 0.34 & 0.38 & 0.35 & 0.36 & 0.15 & 0.32 & 0.29 & 0.31 & 0.33 & 0.41 & 0.53 & 0.51\\ \cmidrule(ll){2-18}
				&\textbf{SA}& 0.41 & 0.57 & 0.59 & 0.67 & 0.27 & 0.42 & 0.39 & 0.44 & 0.31 & 0.44 & 0.45 & 0.46 & 0.32 & 0.42 & 0.49 & 0.51\\    
				\bottomrule
		\end{tabular}}
		\vspace{-0.1in}
		\caption{The improvement of generated video quality in different categories of physical rules in the VideoPhy prompt dataset}	\label{tab:diff_categories_videophy}
	\end{table*}
	
	\begin{table*}[ht]
		\centering
		\vspace{-0.1in}
		{\fontsize{7.5}{9}\selectfont
			\begin{tabular}{cc||cccc||cccc||cccc||cccc}
				\toprule
				\multicolumn{2}{c||}{} & \multicolumn{4}{c||}{\textbf{CogVideoX-5B}} & \multicolumn{4}{c||}{\textbf{CogVideoX-2B}} & \multicolumn{4}{c||}{\textbf{OpenSora}} & \multicolumn{4}{c}{\textbf{VideoCrafter}}\\
				\midrule[1pt]
				\multicolumn{2}{c||}{
					\textbf{Round}} & \textbf{1} & \textbf{2} &\textbf{3} & \textbf{4} &\textbf{1} & \textbf{2} &\textbf{3} & \textbf{4} & \textbf{1} & \textbf{2} &\textbf{3} & \textbf{4} & \textbf{1} & \textbf{2} &\textbf{3} & \textbf{4}\\
				\midrule[1pt]
				\multirow{2}{*}{\textbf{Mechanics}}
				&\textbf{PC}&0.19&0.25&0.34 & 0.35 & 0.12 & 0.16 & 0.18 & 0.24 & 0.11 & 0.13 & 0.17 & 0.22 & 0.14 & 0.23 & 0.29 & 0.28\\ \cmidrule(ll){2-18}
				&\textbf{SA}&0.21&0.28&0.29 & 0.32 & 0.11 & 0.18 & 0.19 & 0.22 & 0.19 & 0.21 & 0.27 & 0.32 & 0.20 & 0.24 & 0.28 & 0.35\\\midrule[1pt]	
				\multirow{2}{*}{\textbf{Optics}}
				&\textbf{PC}&0.22&0.35&0.41& 0.39& 0.22 & 0.25 & 0.29 & 0.28 & 0.24 & 0.26 & 0.25 & 0.25 & 0.22 & 0.21 & 0.27 & 0.32\\ \cmidrule(ll){2-18}
				&\textbf{SA}&0.27&0.42&0.39& 0.44& 0.23 & 0.34 & 0.37 & 0.39 & 0.26 & 0.31 & 0.29 & 0.30 & 0.22 & 0.28 & 0.35 & 0.39\\  \midrule[1pt]  
				\multirow{2}{*}{\textbf{Thermal}}
				&\textbf{PC}&0.33&0.35&0.35& 0.35& 0.13 & 0.15 & 0.15 & 0.14 & 0.27 & 0.30 & 0.31 & 0.33 & 0.25 & 0.28 & 0.26 & 0.28\\ \cmidrule(ll){2-18}
				&\textbf{SA}&0.22&0.36&0.43& 0.45& 0.12 & 0.16 & 0.24 & 0.27 & 0.23 & 0.25 & 0.37 & 0.36 & 0.25 & 0.37 & 0.41 & 0.43\\  
				\bottomrule
		\end{tabular}}
		\vspace{-0.1in}
		\caption{The improvement of generated video quality in different categories of physical rules in the PhyGenBench prompt dataset}	\label{tab:diff_categories_phygenbench}
		\vspace{-0.2in}
	\end{table*}
	
\begin{figure}[ht]
		\centering
		\vspace{-0.1in}
		\includegraphics[width=\linewidth]{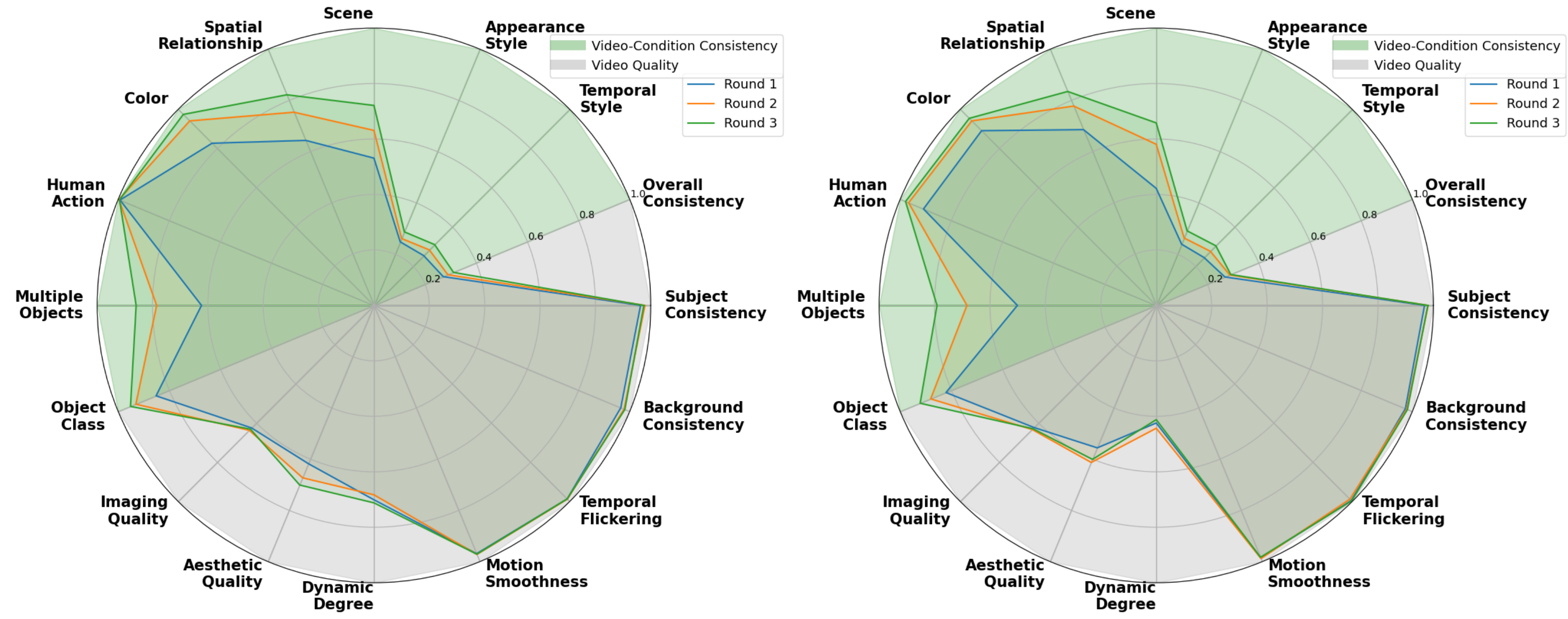} 
		\vspace{-0.2in}
		\caption{PhyT2V VBench evaluation results with CogVideoX-5B (left) and OpenSora (right)}
		\vspace{-0.15in}
		\label{fig:vbench}
	\end{figure}

	Our evaluation results on the VBench metrics are shown in Figure \ref{fig:vbench}, where numbers in Round 1 are the T2V model's original performance in current VBench leaderboard, and iterative prompt refinements by PhyT2V in Round 2 \& 3 noticeably improve the performance in many dimensions. In particular, large improvements are noted in most dimensions of Video-Condition Consistency, showing that PhyT2V improves T2V model's adherence to prompts and real-world physical rules underlying the prompts.
	
	\subsection{Different Domains of Physical Rules}
	\label{subsec:experiment_categories}
	\vspace{-0.05in}
	We also conducted in-depth analysis on PhyT2V's performance on improving the generated video's quality in different domains of real-world physical rules, using the CogVideoX-5B as the T2V model and ChatGPT 4 as the prompt enhancer. As shown in Table \ref{tab:diff_categories_videophy} and \ref{tab:diff_categories_phygenbench}, PhyT2V achieves large improvements in most domains of physical rules. Especially in domains where physical interaction between objects are more subtle and difficult to be precisely captured, such as interaction with fluids and thermal-related scene changes, such improvements will be even higher. 
	
	\begin{figure}
		\centering
		\vspace{-0.1in}
		\includegraphics[width=\linewidth]{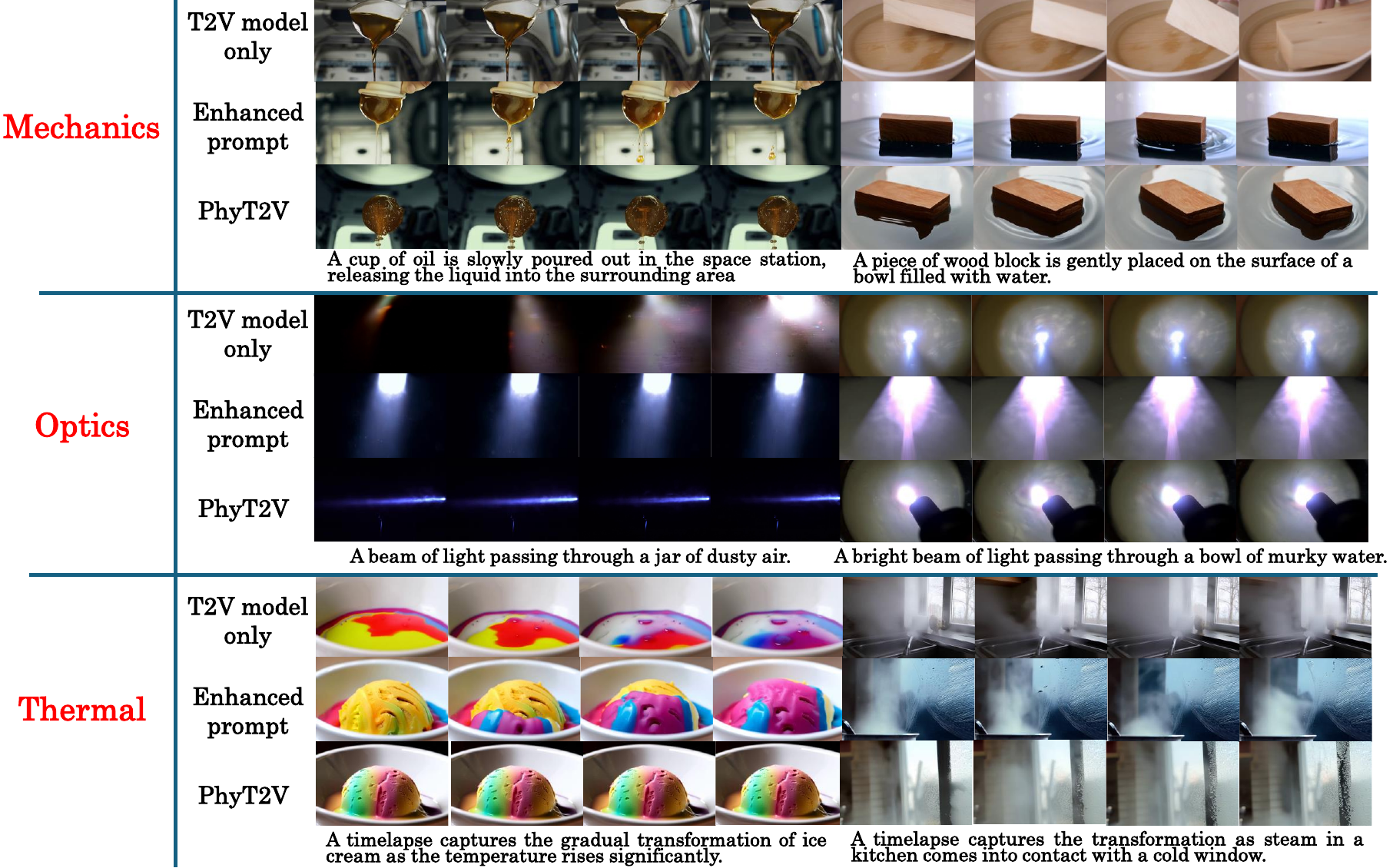}
		\vspace{-0.2in}
		\caption{Examples of videos generated using different categories of prompts in the PhyGenBench dataset}
		\label{fig:example_phygenbench}
	\end{figure}
	
	\begin{figure}
		\centering
		\vspace{-0.15in}
		\includegraphics[width=\linewidth]{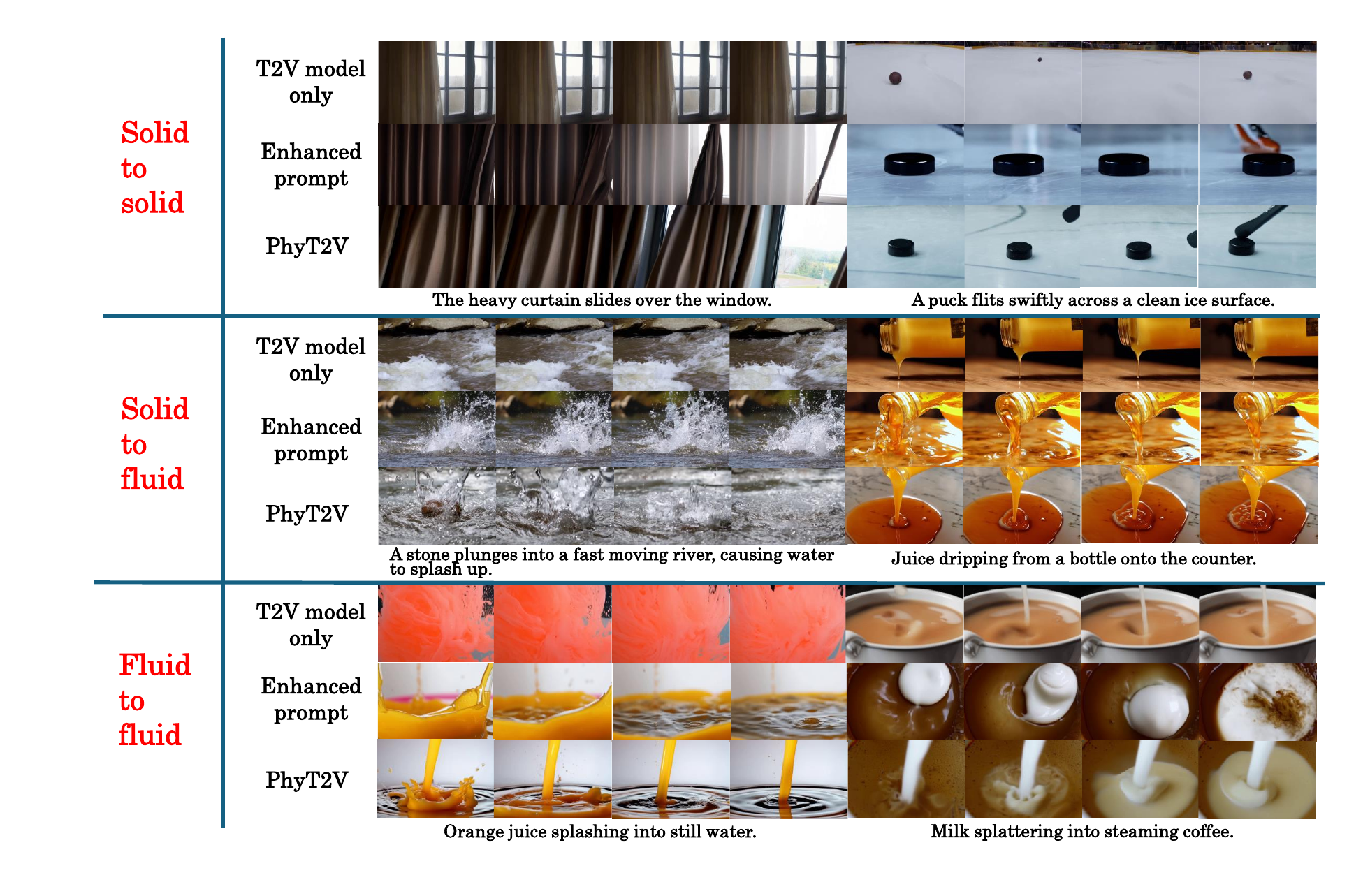}
		\vspace{-0.2in}
		\caption{Examples of videos generated using different categories of prompts in the VideoPhy dataset}
		\label{fig:example_videophy}
		\vspace{-0.15in}
	\end{figure}
	
	These improvements are also exemplified with sample videos and their related input prompts in Figure \ref{fig:example_videophy} and Figure \ref{fig:example_phygenbench}. With LLM reasoning and iterative refinement, PhyT2V can largely enhance the T2V model's capability when encountering out-of-distribution prompts, by providing correct and sufficient contexts to ensure that the T2V model's video generation correctly capture the key objects and interaction between objects. For example, when the prompt of ``juice dropping from a bottle onto the counter'', PhyT2V correctly depicts the juice's slow diffusion on the counter. 
	More examples can be found in Appendix B.
	
	\vspace{-0.05in}
	\subsection{Ablation Study}
	\vspace{-0.05in}
	\label{subsec:ablation}
	We conduct an ablation study to evaluate the necessity of both the physical rule reasoning (Step 1) and the mismatch reasoning (Step 2) within our PhyT2V workflow, by removing one of these steps from the refinement process to assess its impact on physical-grounded video generation.
	
\begin{figure}[ht]
		\centering
		 \vspace{-0.1in}
		\includegraphics[width=\linewidth]{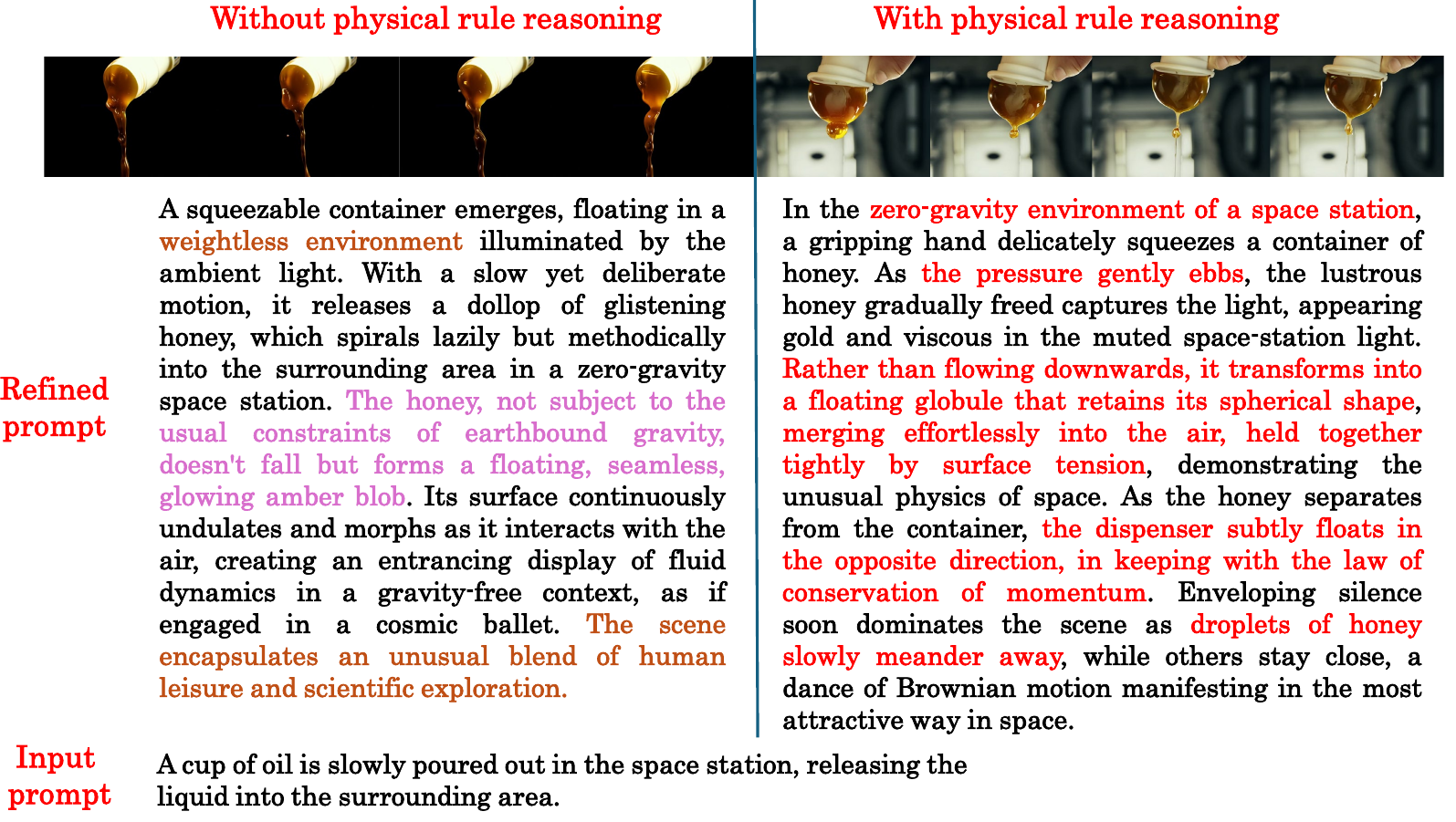}
		\vspace{-0.25in}
		\caption{Ablation study on Step 1 of physical rule reasoning}
			\vspace{-0.1in}
		\label{fig:ablation_step1}
	\end{figure}

	\noindent\textbf{Physical rule reasoning (Step 1).} As shown in Figure \ref{fig:ablation_step1}, the Step 1 of physical rule reasoning significantly enhances the T2V process by providing a more detailed and coherent description of the principal object's physical status, such as motion, states and deformation (red texts in Figure \ref{fig:ablation_step1}), all grounded in relevant physical laws. By anchoring the prompt in established physical rules, this step also help avoid unnecessary texts (brown texts in Figure \ref{fig:ablation_step1}) and vague physical rule descriptions (purple texts in Figure \ref{fig:ablation_step1}), hence achieving a higher PC score.
	
	\begin{figure}[ht]
		\centering
		\includegraphics[width=\linewidth]{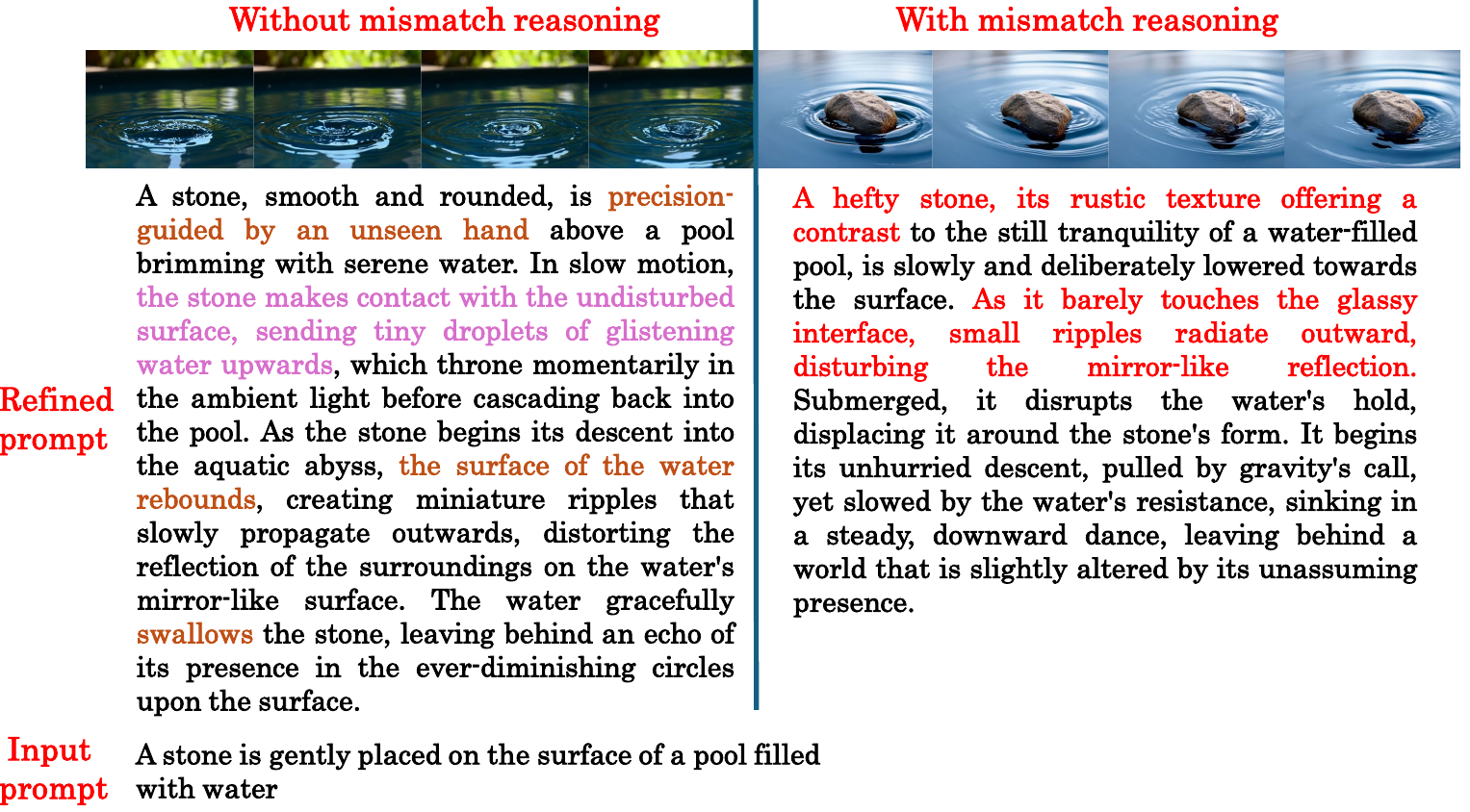}
		\vspace{-0.25in}
		\caption{Ablation study on Step 2 of mismatch reasoning}
		\label{fig:ablation_step2}
		\vspace{-0.25in}
	\end{figure}
	
	\noindent\textbf{Mismatch reasoning (Step 2).} The Step 2 of mismatch reasoning addresses details that may have been overlooked in the previous iteration of the generated video as shown in Figure \ref{fig:ablation_step2}. This step plays a critical role in the iterative refinement process by identifying and correcting discrepancies between expected and observed outputs. By enhancing the model's focus on the principal object, the mismatch reasoning step reduces the likelihood of losing attention to important features (brown and purple texts in Figure \ref{fig:ablation_step2}), improving the fidelity and relevance of generated video content (red texts in Figure \ref{fig:ablation_step2}) towards a higher SA score.
	
	Overall, our study shows that both reasoning steps are integral to PhyT2V, contributing to a more robust and semantically-aligned generation of refined prompts in Step 3. More detailed ablation studies are in Appendix C.

	
	

	
	
	
	
	
	
	\vspace{-0.1in}
	\section{Conclusion}
	\vspace{-0.05in}
	In this paper, we present PhyT2V, a novel data-independent T2V generation framework designed to enhance the generalization capability of existing T2V models to out-of-distribution domains. By incorporating CoT reasoning and step-back prompting, PhyT2V systematically refines T2V prompts to ensure adherence to real-world physical principles without necessitating additional model retraining or reliance on additional conditions. 
	Evaluation results indicate that PhyT2V achieves a 2.3x enhancement in physical realism compared to baseline T2V models and outperforms state-of-the-art T2V prompt enhancers by 35\%. 
	
	\vspace{-0.05in}
	\section*{Acknowledgments}
	\vspace{-0.05in}
	We thank the reviewers and the area chair for their insightfulcomments and feedback. This work was supported in part by National Science Foundation (NSF) under grant number IIS-2205360, CCF-2217003, CCF-2215042, and National Institutes of Health (NIH) under grant number R01HL170368.

	
	
	
	
	
	
	
	

	
	{
		\small
		\bibliographystyle{ieeenat_fullname}
		\bibliography{cvpr2025}
	}

\newpage
\pagebreak
\appendix

\section{Details of Evaluation Setup}

Since our proposed technique of PhyT2V does not involve any efforts of retraining the T2V model, in this section we describe details about our evaluation setup of the LLM inference for CoT and step-back reasoning. 

In our evaluations, we use 4 T2V generation models, including CogVideoX-5B \cite{yang2024cogvideoxtexttovideodiffusionmodels}, CogVideoX-2B \cite{yang2024cogvideoxtexttovideodiffusionmodels}, OpenSora \cite{opensora} and VideoCrafter \cite{chen2024videocrafter2}. We choose to use these models as they are all built with transformer based diffusion model, which enhanced the semantic adherence by using the cross attention mechanism, and hold a high score on both the VideoPhy dataset and PhyGenBench dataset leader board. We use Tarsier \cite{wang2024tarsier} as the video captioning model for it achieves state of art on multiple video question answering datasets, which ensure the precise and detail of the video captioning step in our approach.

Since PhyT2V improves the quality of generated videos through text prompt refinement, we use the Promptist \cite{li2024promptist} and GPT-4o \cite{yang2024worldgpt} as the prompt enhancers, with the same model hyper-parameter setting for the baselines to maintain the consistency between the baseline and our approach. The diffusion model generated video length is setted as 6 second and 8 frames per second with frame resolution 720 $\times$ 480.

To fit the maximum token input length of the T2V model, we limit the word length of the refined prompts to 120, by instructing the ChatGPT4 o1-preview model that is used as the LLM for reasoning. To formatting the output and storage in our approach, the ChatGPT4 o1-preview model are instructed to output in JSON format and output results in each step are saved in a CSV file by row. The ChatGPT4 o1-preview is called by using the API with the input prompt separated into system prompt and user prompt to identify the instruction part and the rest of the original prompt for ChatGPT. The first and second step output is embedded to the third step prompt template by replacing the pre-defined place holder. 

\section{More Examples of Physics-Grounded Videos Generated by PhyT2V}

\subsection{Examples of Generated Videos in Different Categories of Physical Rules}
\begin{figure*}[ht]
	\centering
	\vspace{-1.8in}        
	\includegraphics[width=\linewidth]{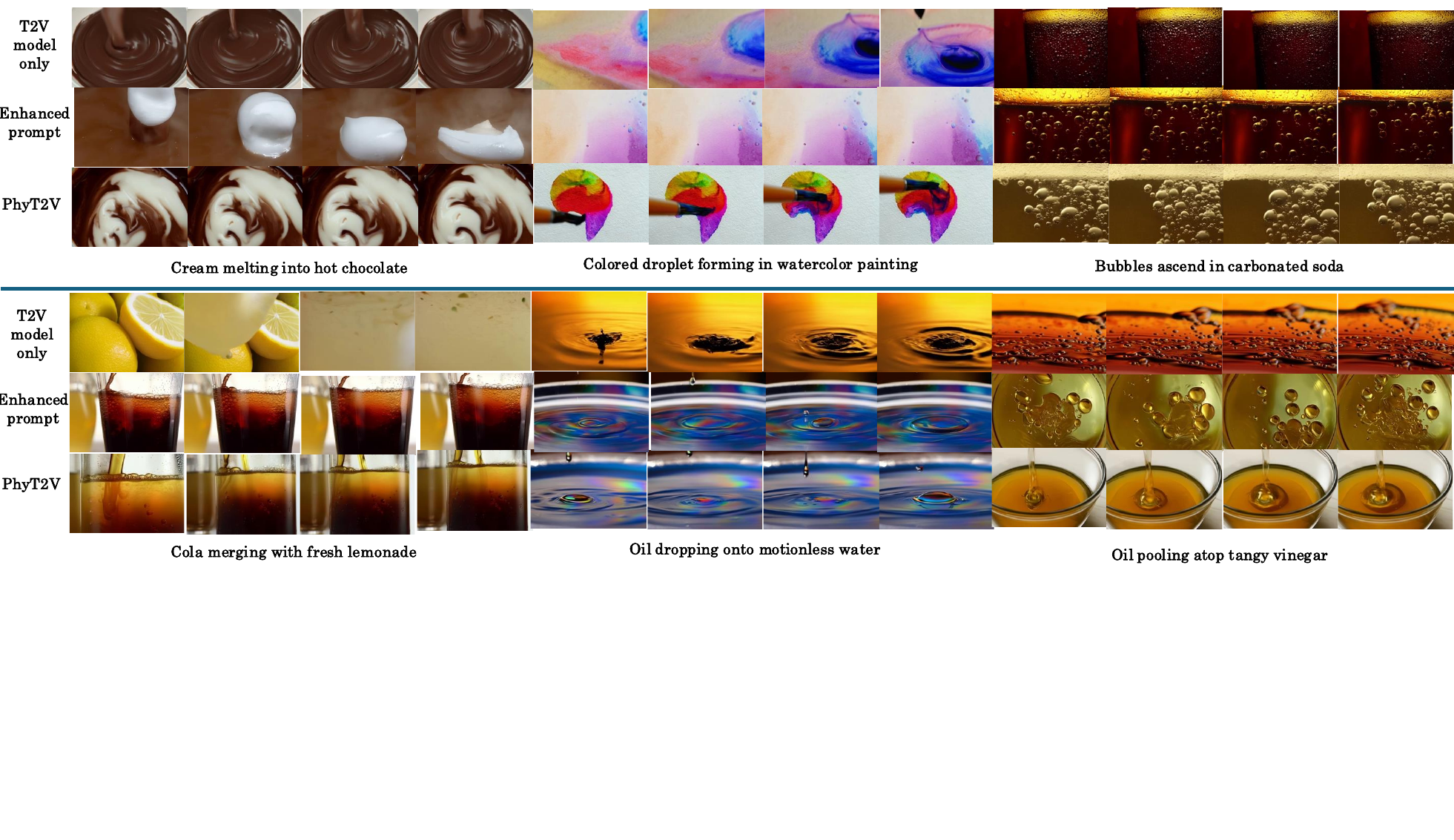} 
	\vspace{-0.3in}
	\caption{Video generation example on fluid to fluid specific prompt in VideoPhy dataset}
	\label{fig:sup_f2f}
\end{figure*}

\begin{figure*}[ht]
	\centering
	\vspace{-0.8in}        
	\includegraphics[width=\linewidth]{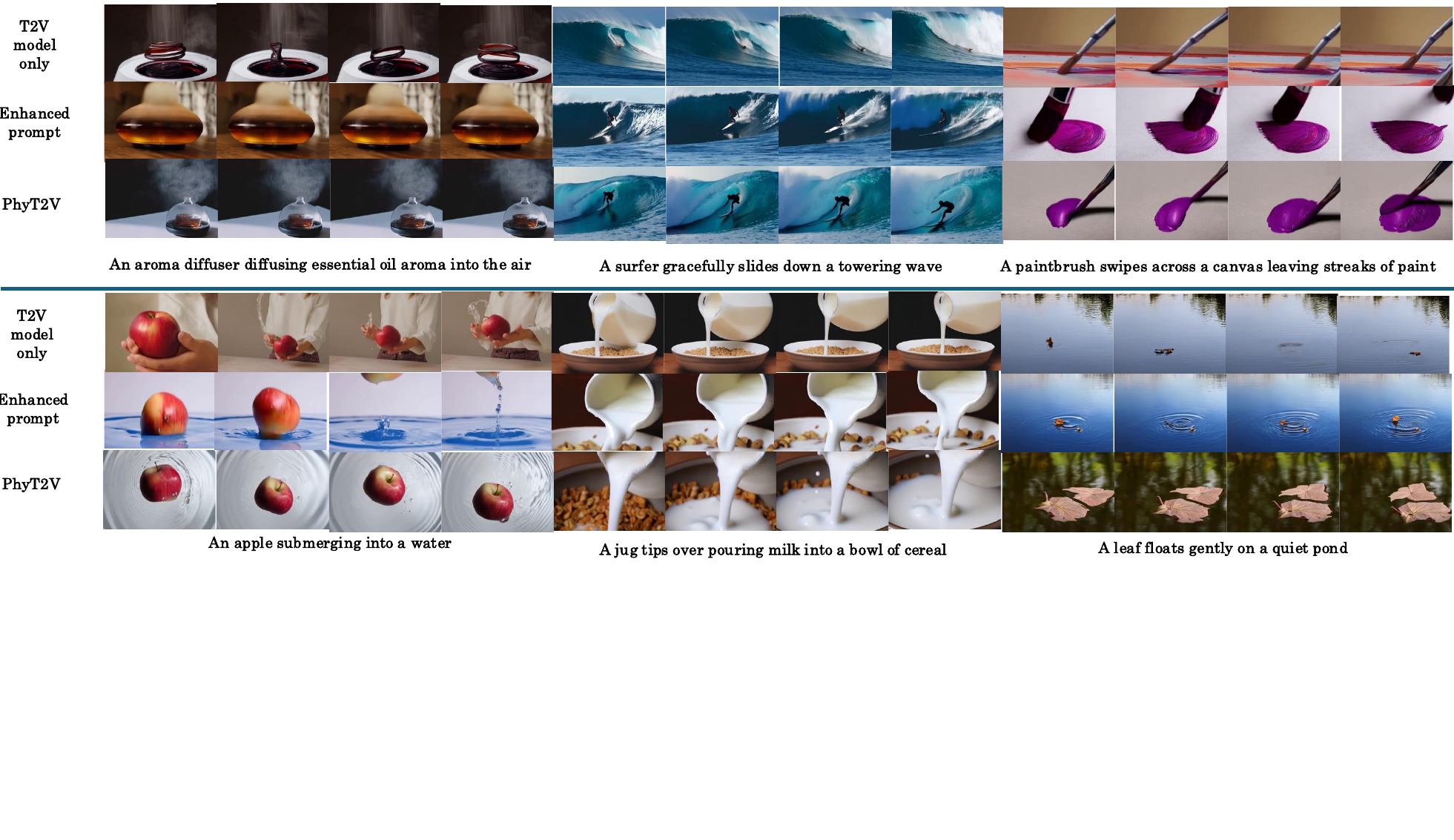} 
	\caption{Video generation example on solid to fluid specific prompt in VideoPhy dataset}
	\label{fig:sup_s2f}
\end{figure*}

\begin{figure*}[ht]
	\centering
	\vspace{-2.8in} 
	\includegraphics[width=\linewidth]{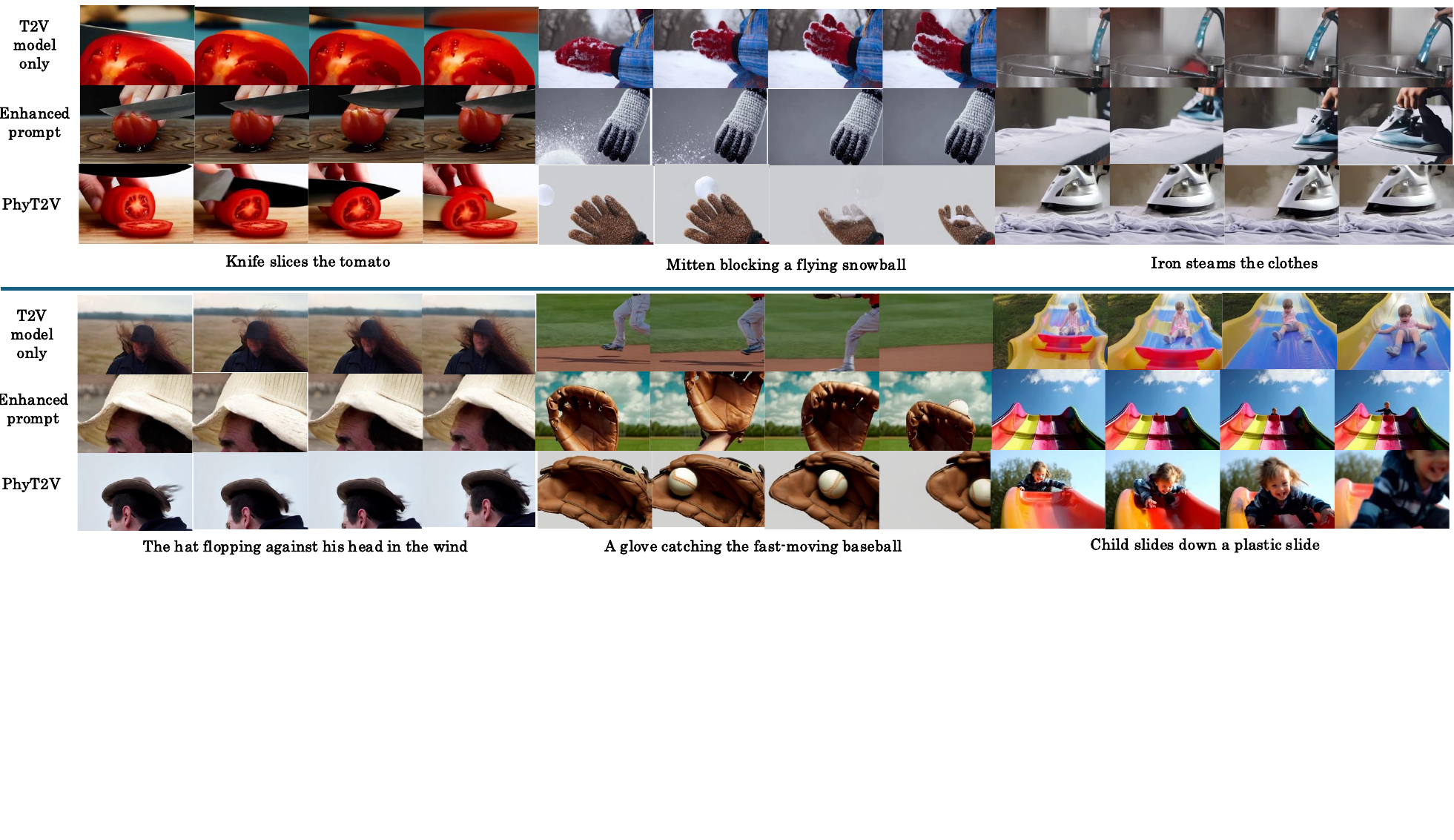} 
	\vspace{-0.3in}
	\caption{Video generation example on solid to solid specific prompt in VideoPhy dataset}
	\label{fig:sup_s2s}
\end{figure*}

\begin{figure*}[ht]
	\centering
	\includegraphics[width=1\linewidth]{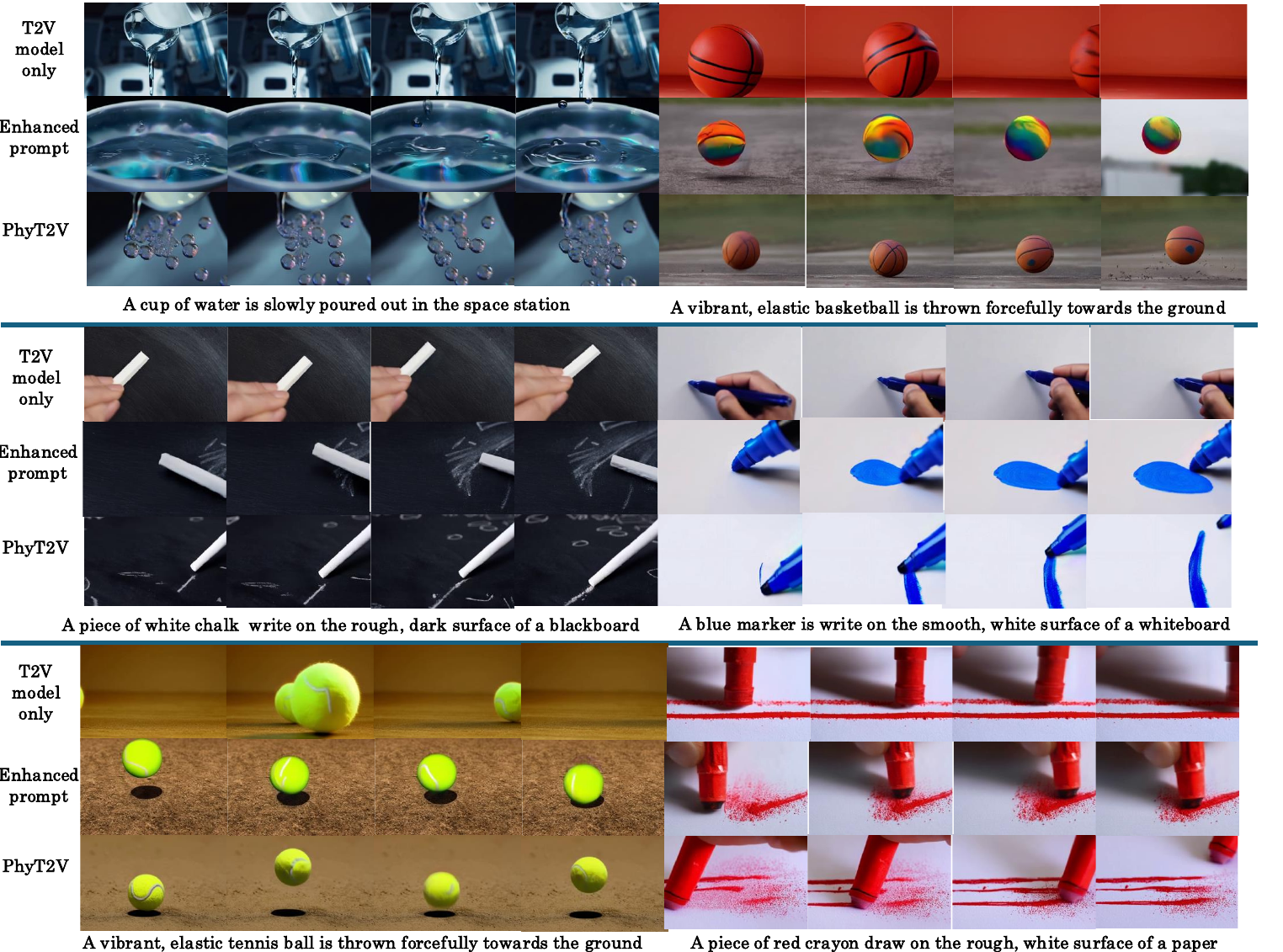} 
	\caption{Video generation example on force specific prompt in PhyGenBench dataset}
	\label{fig:sup_force}
\end{figure*}

\begin{figure*}[ht]
	\centering
	\includegraphics[width=1\linewidth]{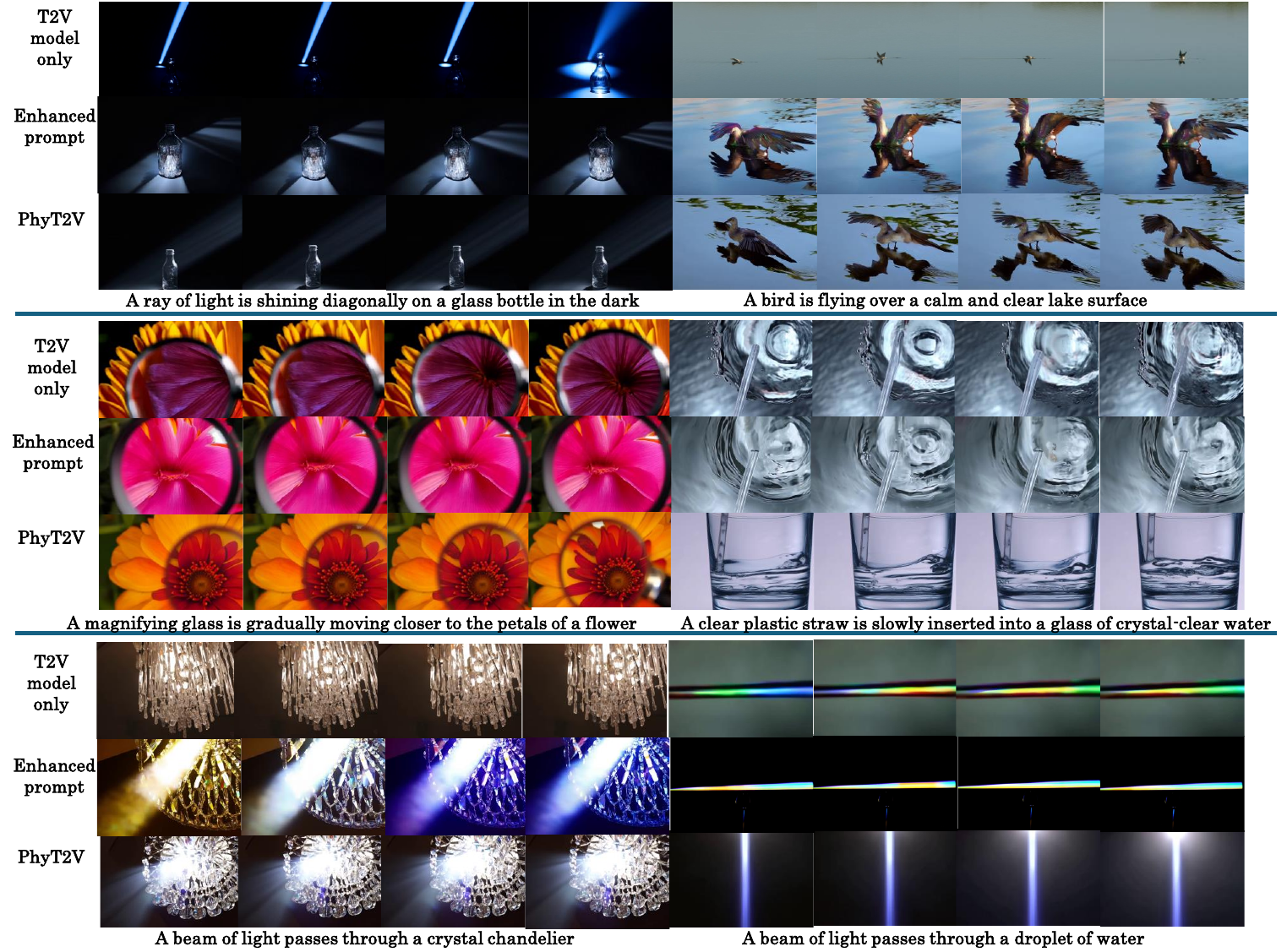} 
	\caption{Video generation example on optics specific prompt in PhyGenBench}
	\label{fig:sup_optics}
\end{figure*}

\begin{figure*}[ht]
	\centering      
	\includegraphics[width=1\linewidth]{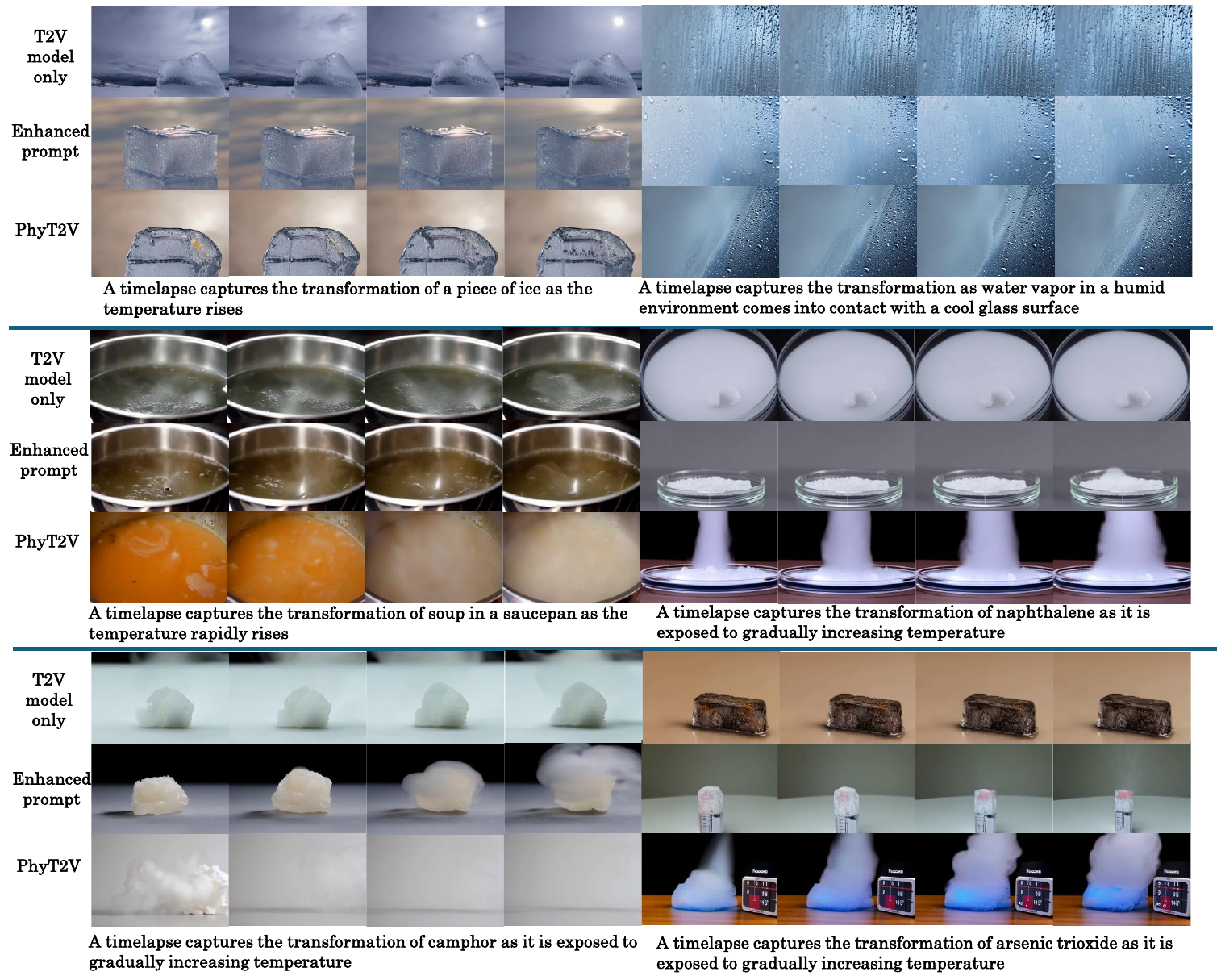} 
	\caption{Video generation example on thermal specific prompt in PhyGenBench}
	\label{fig:sup_thermal}
\end{figure*}

In this subsection, we present the additional comparison examples of generated videos of CogVideo-5b on the VideoPhy and PhyGenBench dataset, Fig \ref{fig:sup_f2f}, \ref{fig:sup_s2f}, \ref{fig:sup_s2s} show the additional comparison of video generation result on fluid to fluid, solid to fluid, solid to solid cases on VideoPhy dataset, Fig \ref{fig:sup_force}, \ref{fig:sup_optics}, \ref{fig:sup_thermal} show the additional comparison of video generation result on force, optics and thermal cases on PhyGenBench dataset.

\subsection{Refinement process details}
\begin{figure*}[ht]
	\centering
	\includegraphics[width=\linewidth]{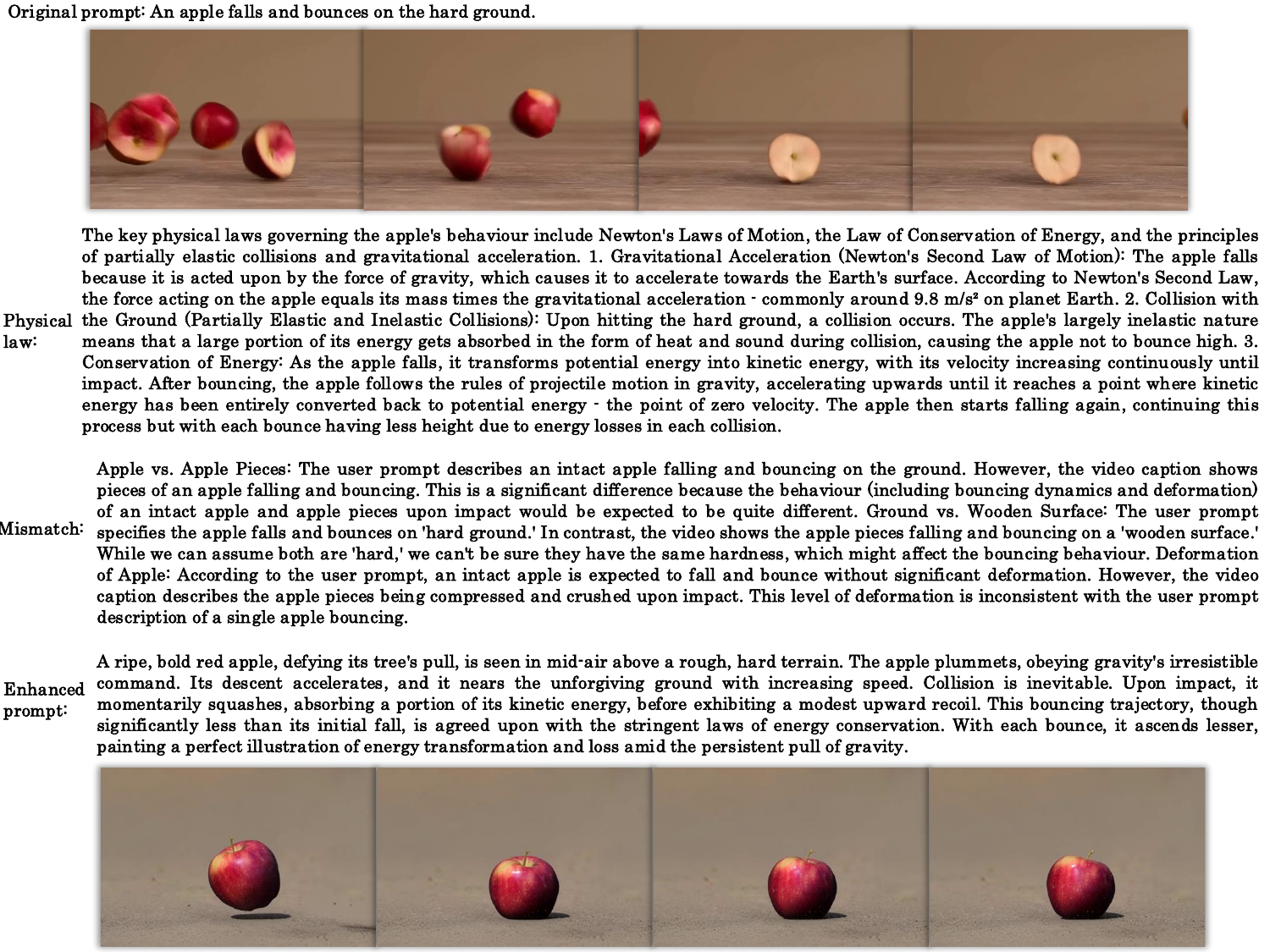} 
	\caption{Refinement detail example on solid to solid specific prompt in VideoPhy dataset}
	\label{fig:sup_prompt_s2s}
\end{figure*}

\begin{figure*}[ht]
	\centering
	\includegraphics[width=\linewidth]{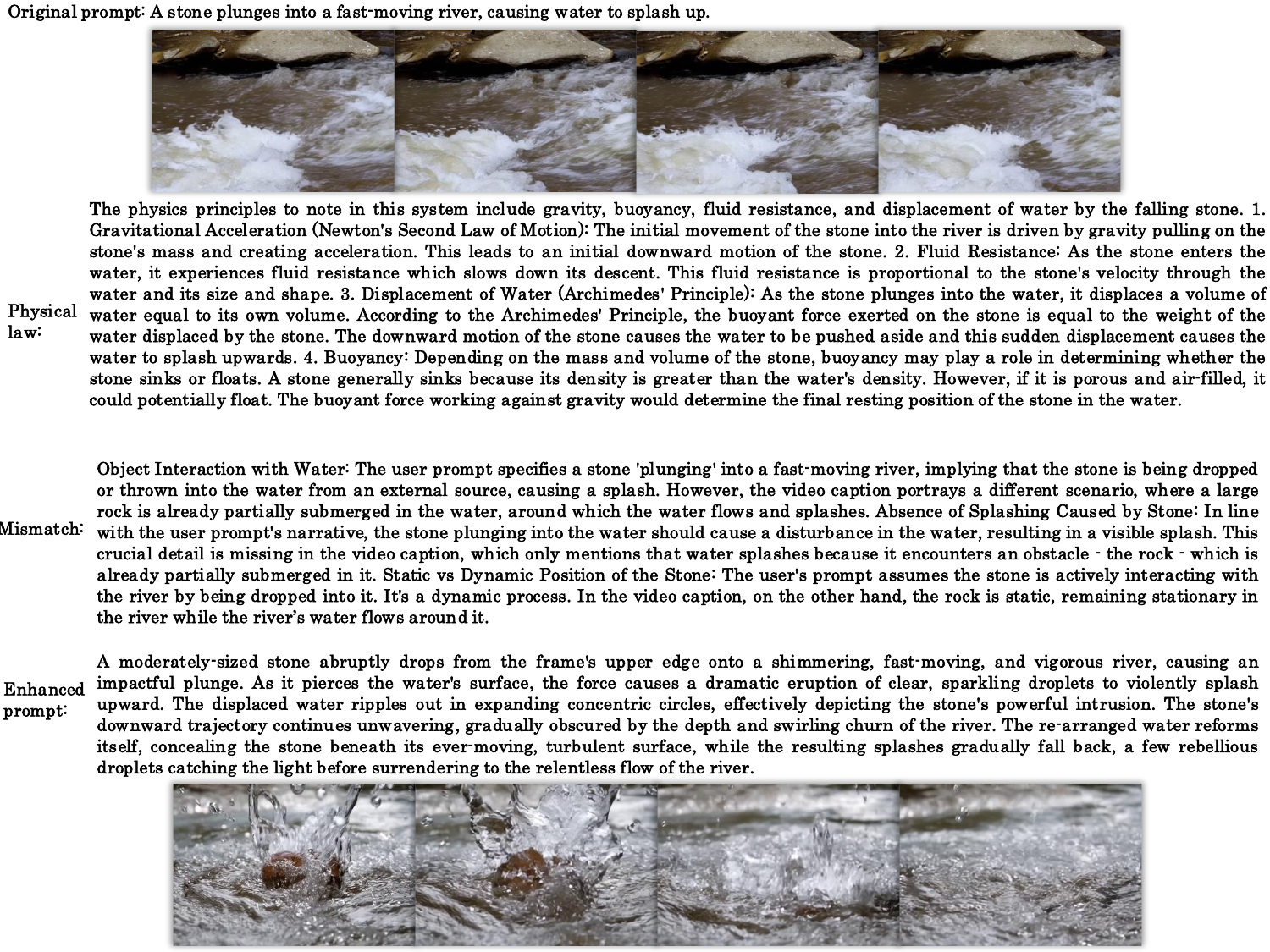} 
	\caption{Refinement detail example on solid to fluid specific prompt in VideoPhy dataset}
	\label{fig:sup_prompt_s2f}
\end{figure*}

\begin{figure*}[ht]
	\centering
	\includegraphics[width=\linewidth]{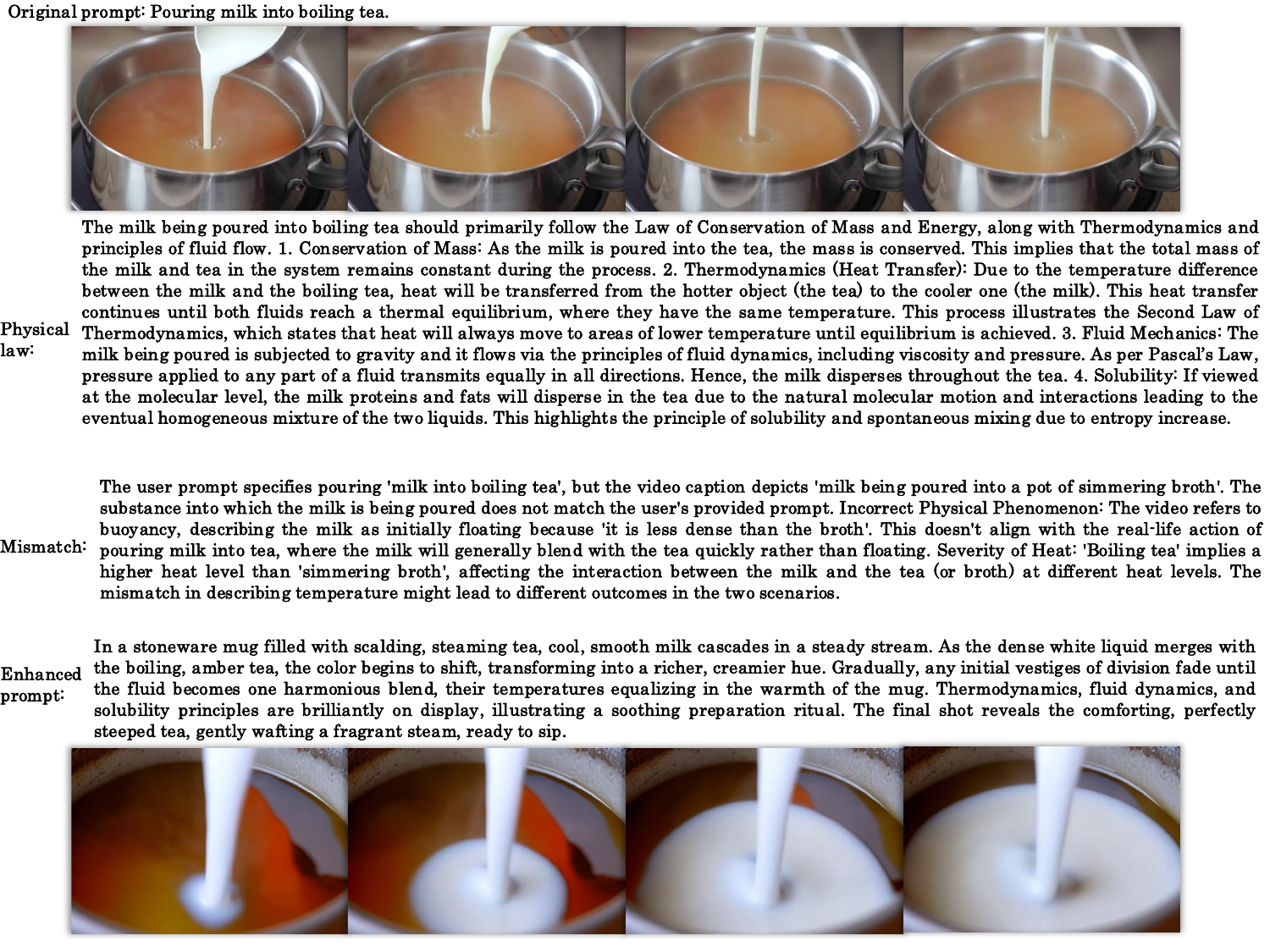} 
	\vspace{-0.3in}
	\caption{Refinement detail example on fluid to fluid specific prompt in VideoPhy dataset}
	\label{fig:sup_prompt_f2f}
\end{figure*}

\begin{figure*}[ht]
	\centering
	\includegraphics[width=\linewidth]{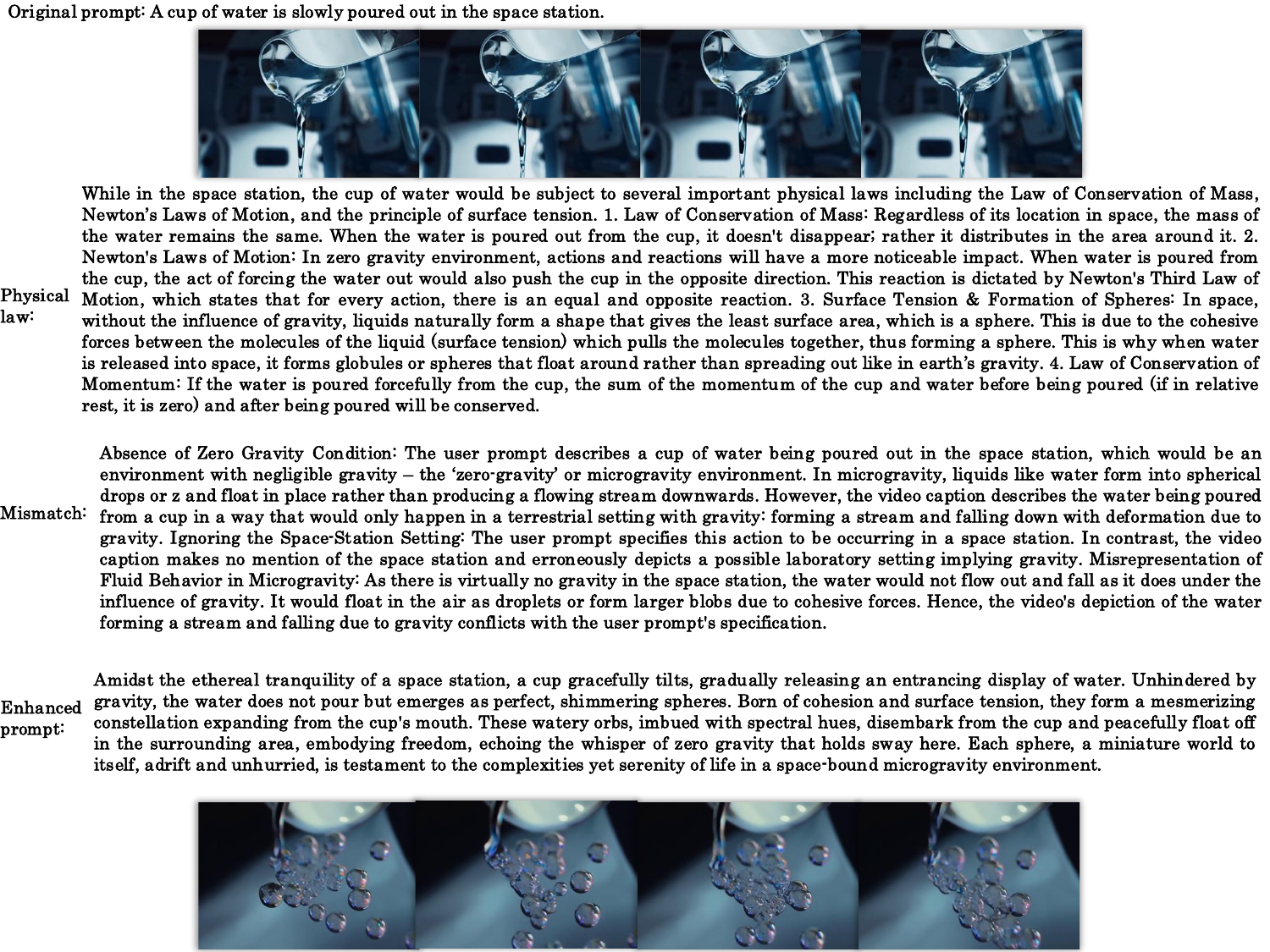} 
	\caption{Refinement detail example on force specific prompt in PhyGenBench dataset}
	\label{fig:sup_prompt_force}
\end{figure*}

\begin{figure*}[ht]
	\centering
	\includegraphics[width=\linewidth]{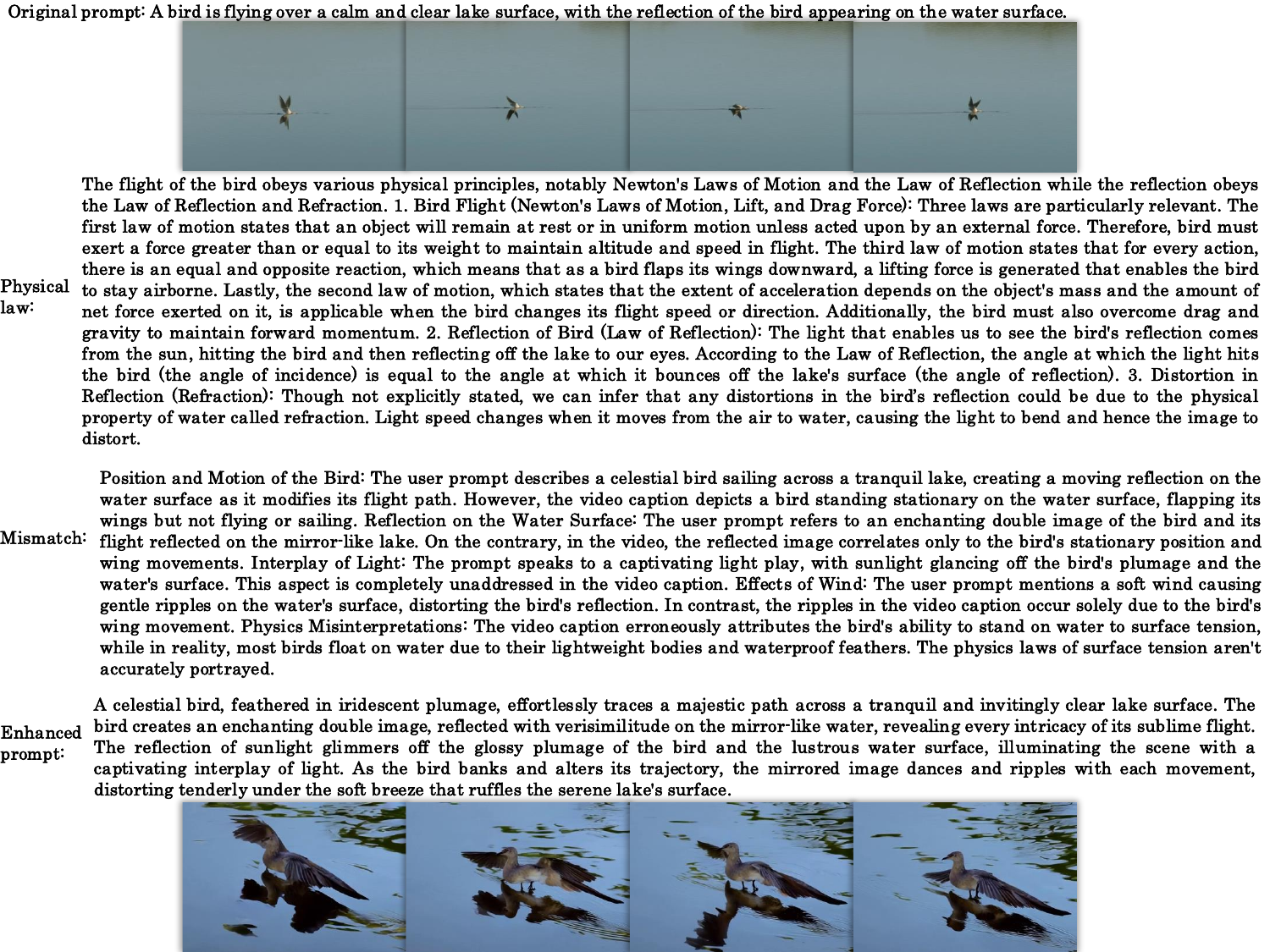} 
	\caption{Refinement detail example on optics specific prompt in PhyGenBench dataset}
	\label{fig:sup_prompt_optics}
\end{figure*}

\begin{figure*}[ht]
	\centering
	\includegraphics[width=\linewidth]{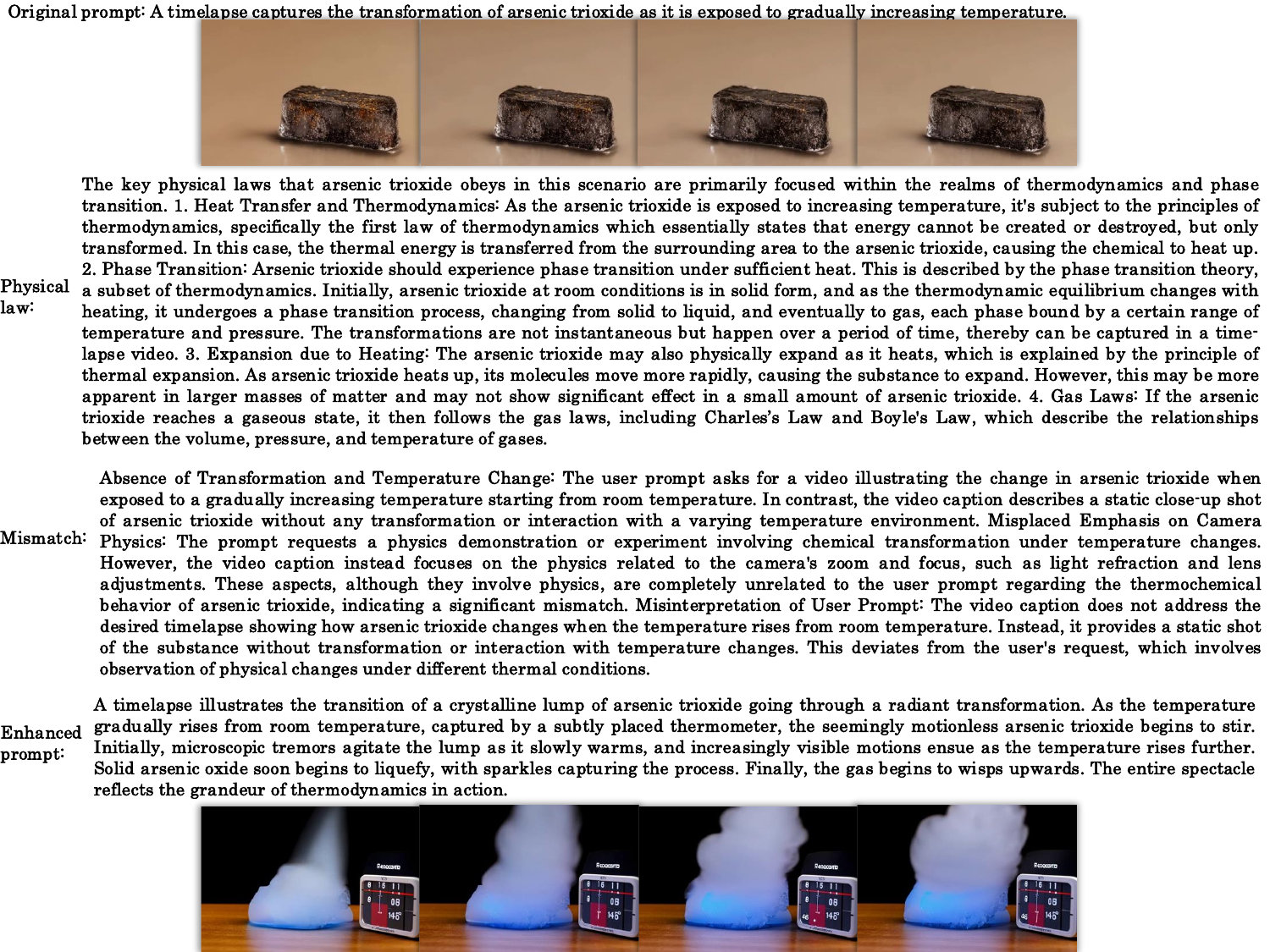} 
	\caption{Refinement detail example on thermal specific prompt in PhyGenBench dataset}
	\label{fig:sup_prompt_thermal}
\end{figure*}

In this subsection, we present the prompt refinement detail by the CogVideo-5b with PhyT2V on the VideoPhy and PhyGenBench dataset, Fig \ref{fig:sup_prompt_f2f}, \ref{fig:sup_prompt_s2f}, \ref{fig:sup_prompt_s2s} show the prompt refinement details of fluid to fluid, solid to fluid, solid to solid cases on VideoPhy dataset, Fig \ref{fig:sup_prompt_force}, \ref{fig:sup_prompt_optics}, \ref{fig:sup_prompt_thermal} show prompt refinement details of  force, optics and thermal cases on PhyGenBench dataset.

\section{Ablation study details}
\subsection{Model size}
\begin{figure*}[ht]
	\centering
	\includegraphics[width=\linewidth]{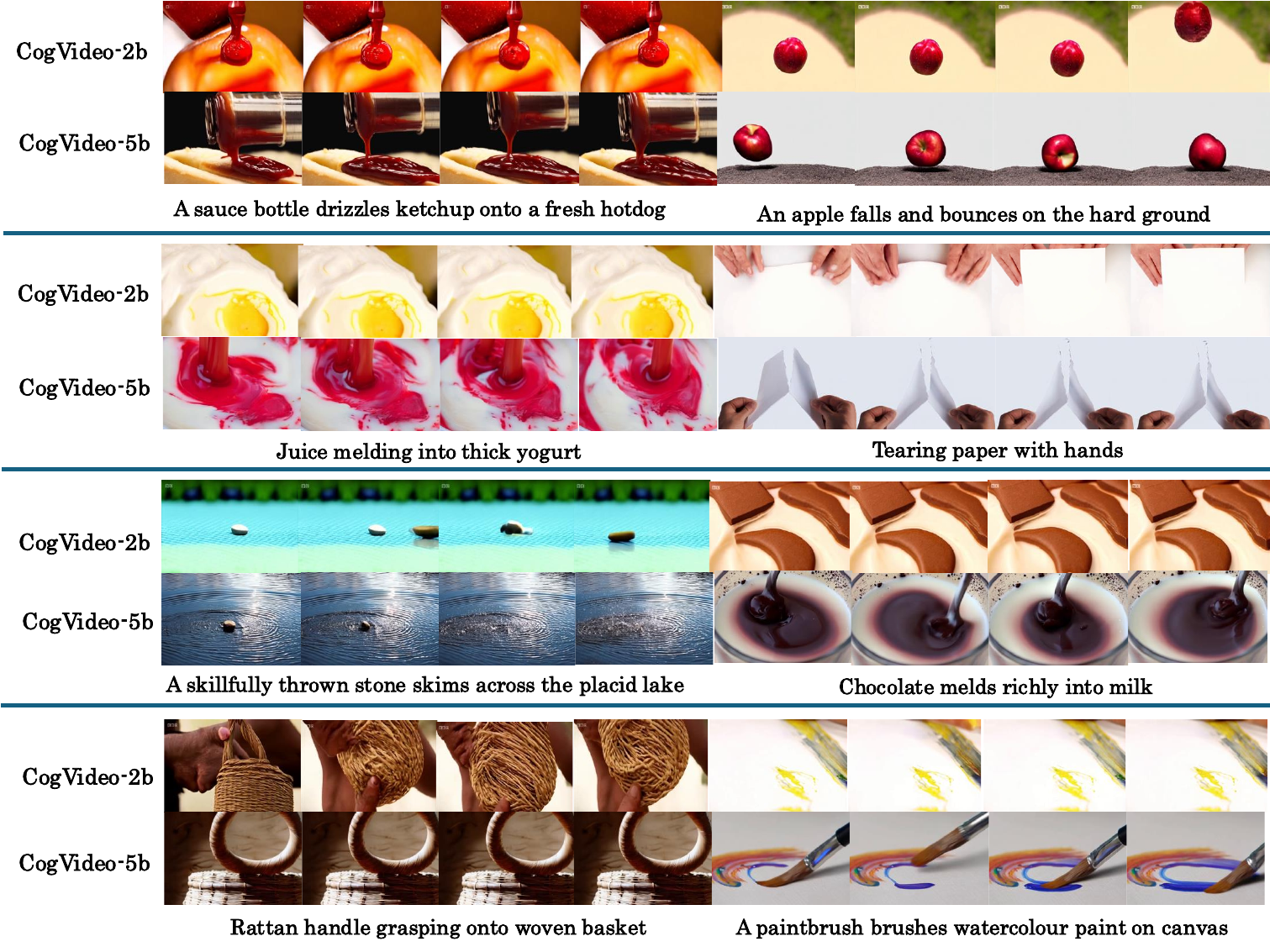}
	\caption{Ablation study on model size}
	\label{fig:ablation_model}
\end{figure*}

We found that the PhyT2V approach can unleashing more power of physical-grounded video generation on a larger model as the result shown by comparing the CogVideo-2b and CogVideo-5b in Figure \ref{fig:ablation_model} .

\subsection{Prompt template component}
\begin{figure*}[ht]
	\centering
	\includegraphics[width=\linewidth]{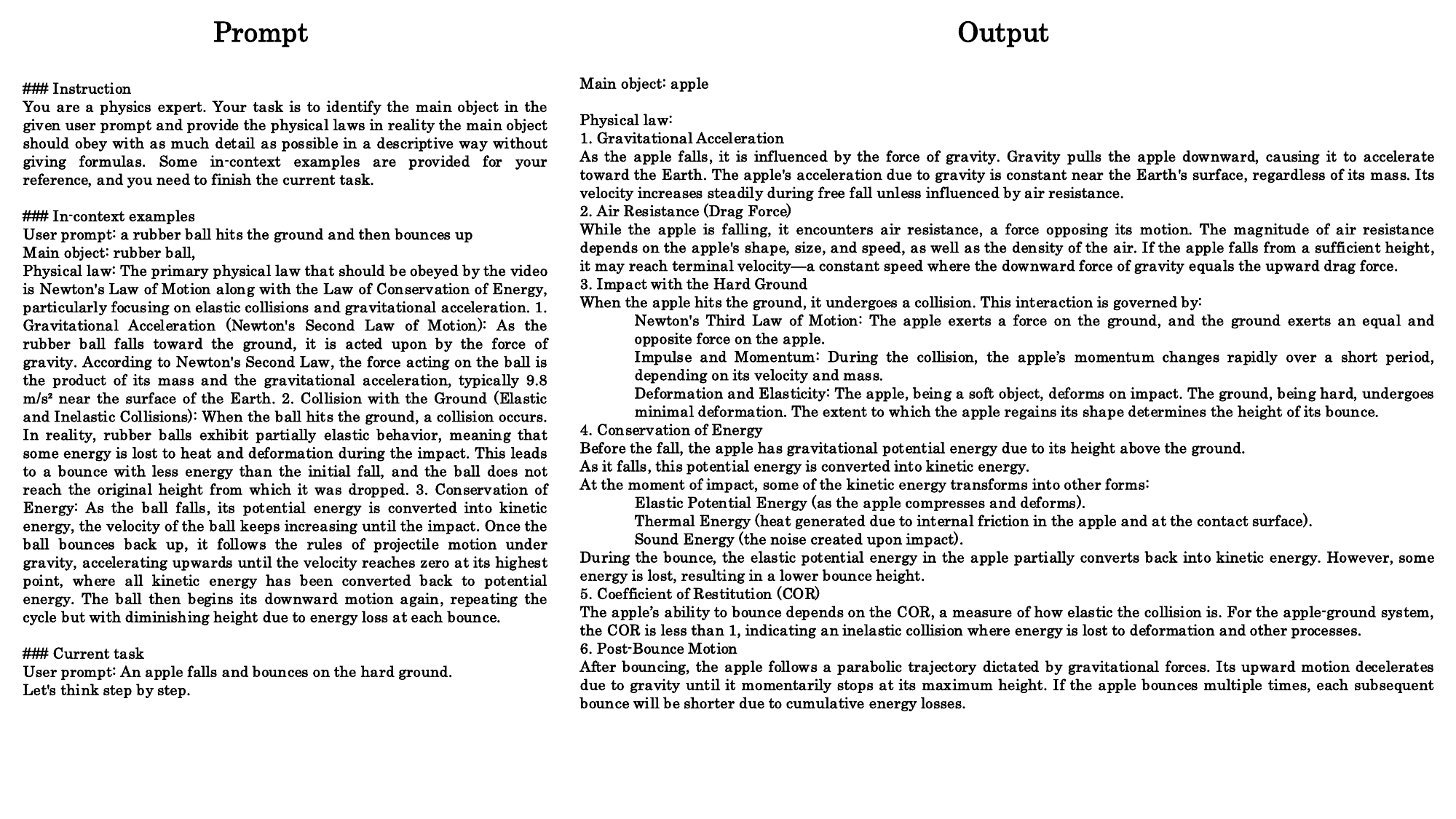} 
	\vspace{-0.7in} 
	\caption{The original prompt and output}
	\label{fig:ablation_prompt_1}
\end{figure*}

\begin{figure*}[ht]
	\centering
	\vspace{-1in}
	\includegraphics[width=\linewidth]{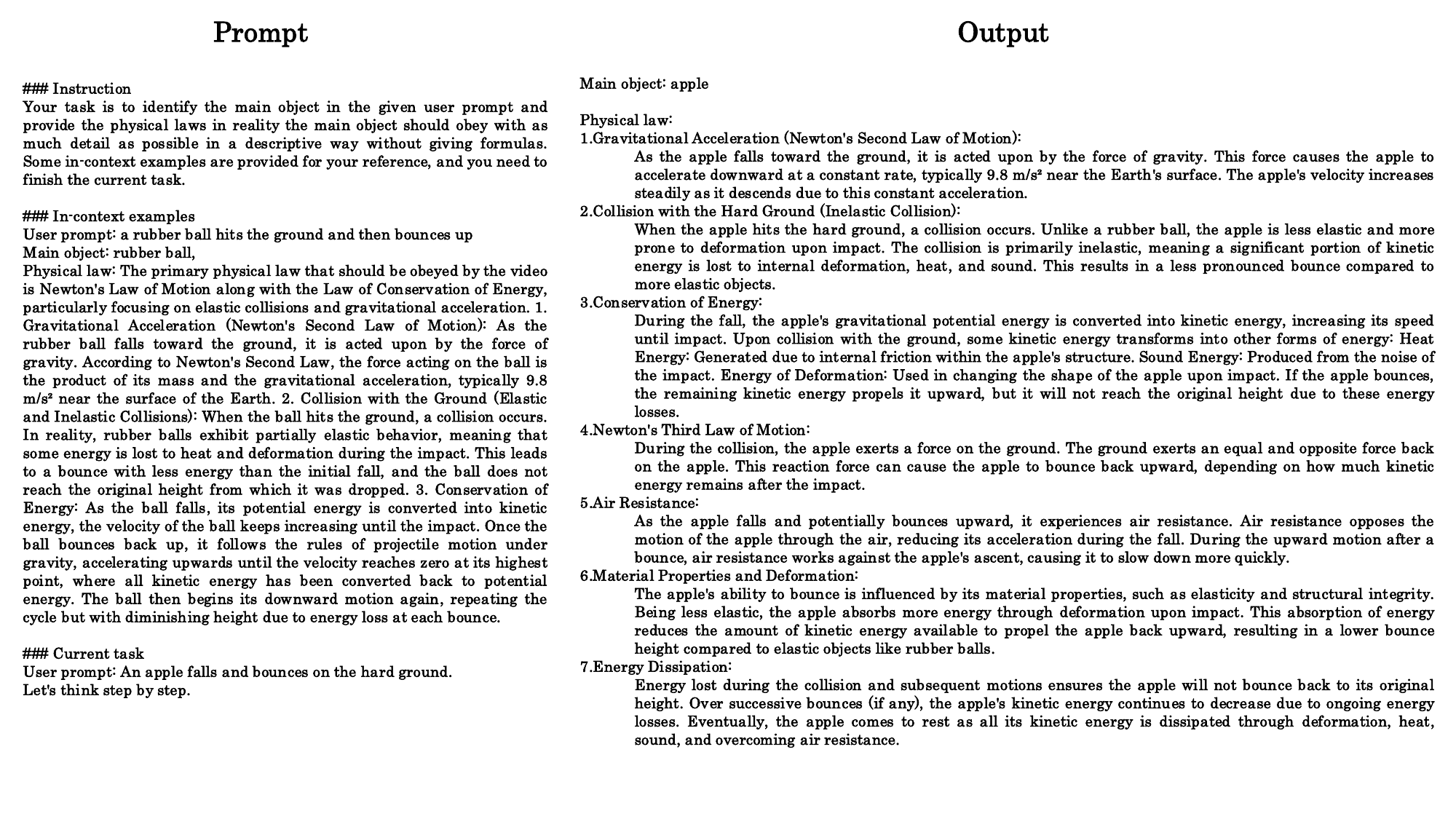} 
	\vspace{-0.5in}
	\caption{The prompt without the role indicator and output}
	\label{fig:ablation_prompt_2}
\end{figure*}

\begin{figure*}[ht]
	\centering
	\includegraphics[width=\linewidth]{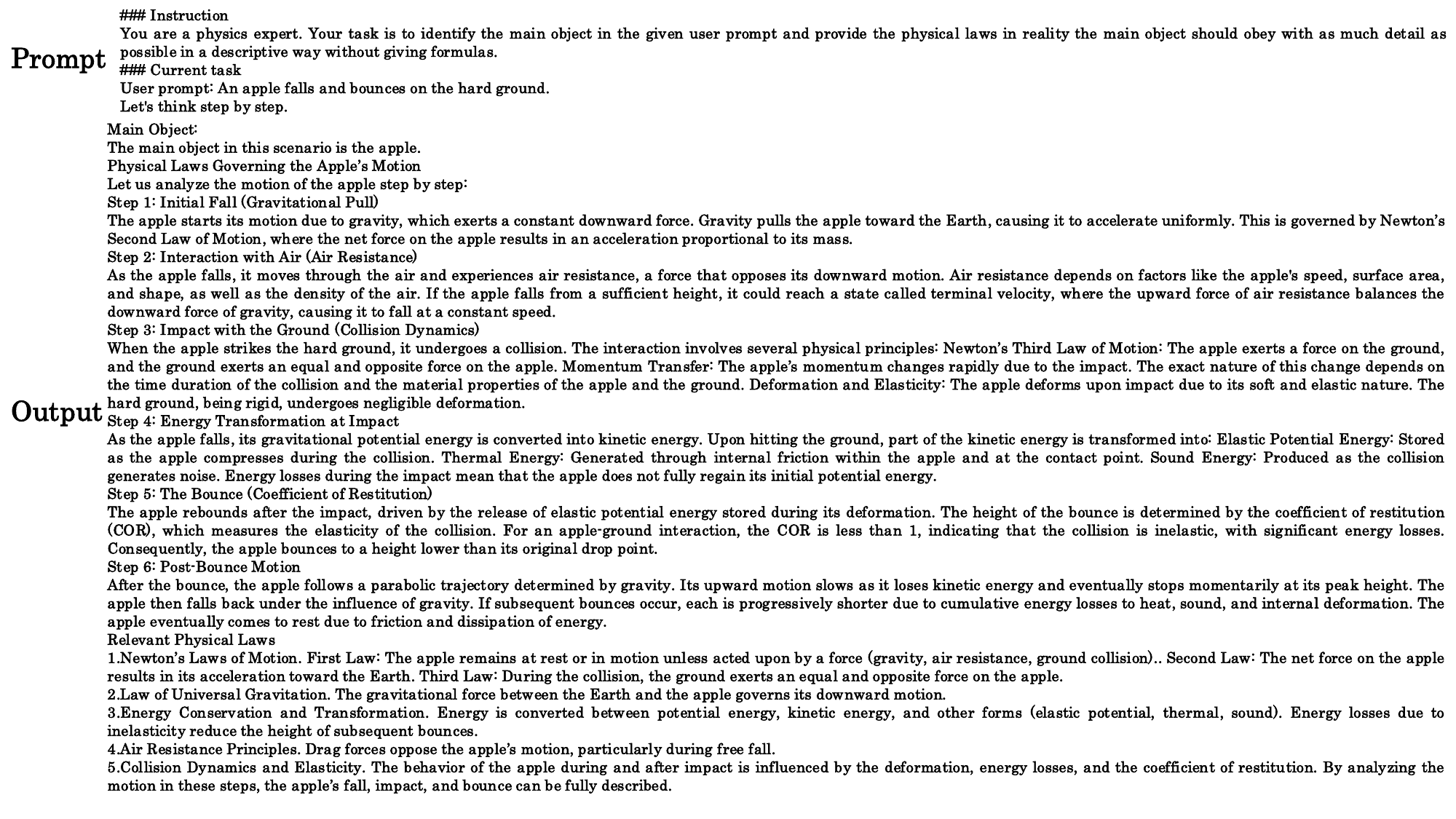} 
	\vspace{-0.3in}
	\caption{The prompt without the in-context example and output}
	\label{fig:ablation_prompt_3}
\end{figure*}

In this section some part of the prompt template component is removed to show the necessity of the corresponding components as shown in Figure\ref{fig:ablation_prompt_1}, \ref{fig:ablation_prompt_2}, \ref{fig:ablation_prompt_3}. Without the role indicator sentence, the generated output content is lake of precise information, without the in-context examples, the GPT can not generated the output in an expected format.

\section{Failure cases}
PhyT2V may be ineffective in two categories of cases. 

First, many T2V models exhibit temporal flickering or inconsistent object trajectories, due to absence of long-term temporal coherence mechanisms in model design. Even if prompts are refined to emphasize smooth temporal transitions or continuous motions, these requirements may not be achieved due to model's limitations. 

Second, T2V models are typically trained on large datasets, which often lack samples of rare or complex physical phenomena. Hence, these models struggle in scenarios that are underrepresented in the training data. Even with highly specific prompts, the T2V models may still fail to extrapolate effectively to these underrepresented cases.

\begin{figure*}[ht]
	\centering
	\includegraphics[width=\linewidth]{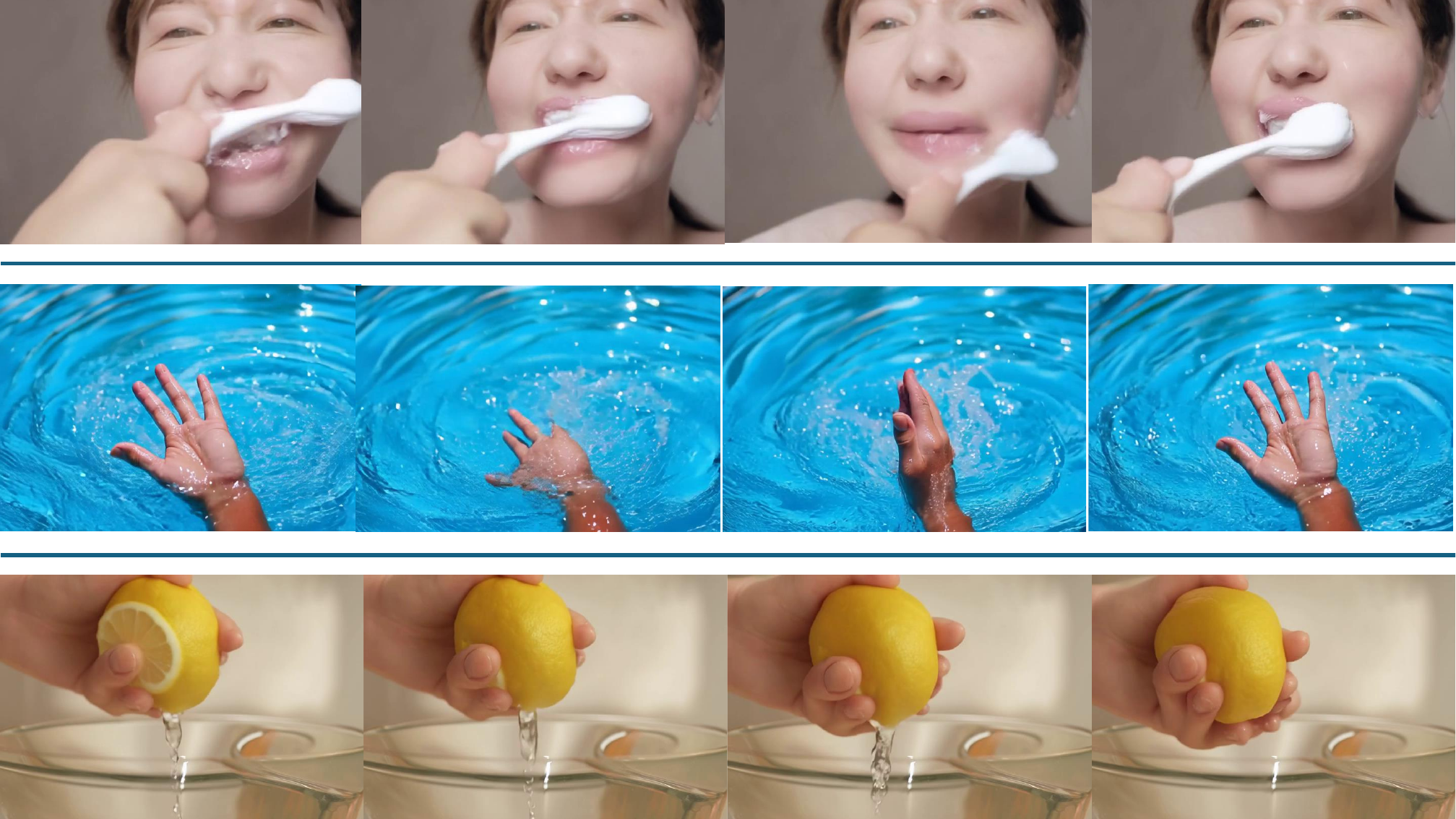} 
	\caption{Failure case example on human hand related video generation}
	\label{fig:fail_case}
\end{figure*}

For some specific generation content categories, we found that even with the PhyT2V refined several rounds, the diffusion model still failed to precisely generating human body, especially on human hands as shown in Figure \ref{fig:fail_case}.

\begin{figure*}[ht]
	\centering    
	\includegraphics[width=\linewidth]{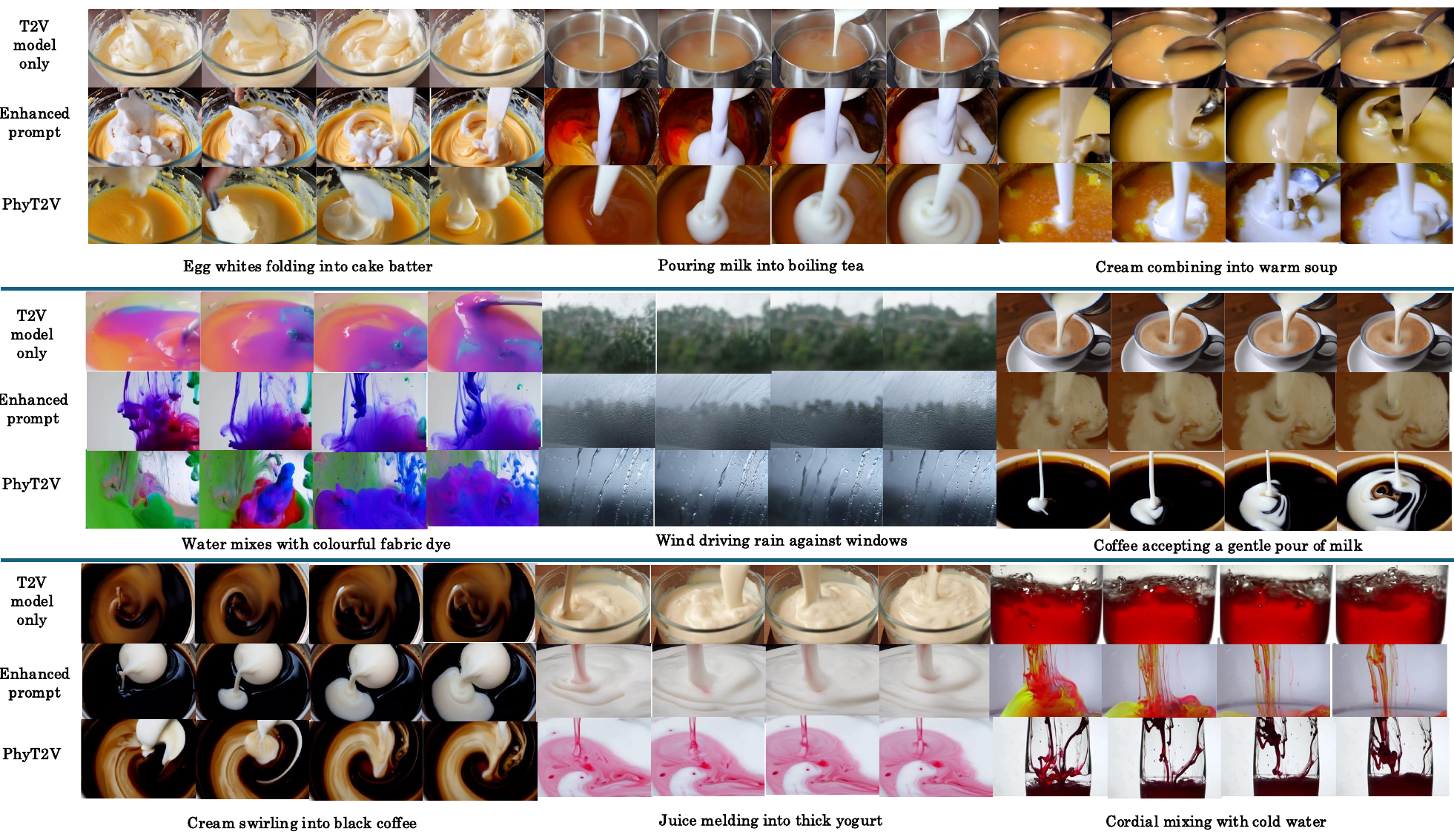} 
\end{figure*}

\begin{figure*}[ht]
	\centering   
	\includegraphics[width=\linewidth]{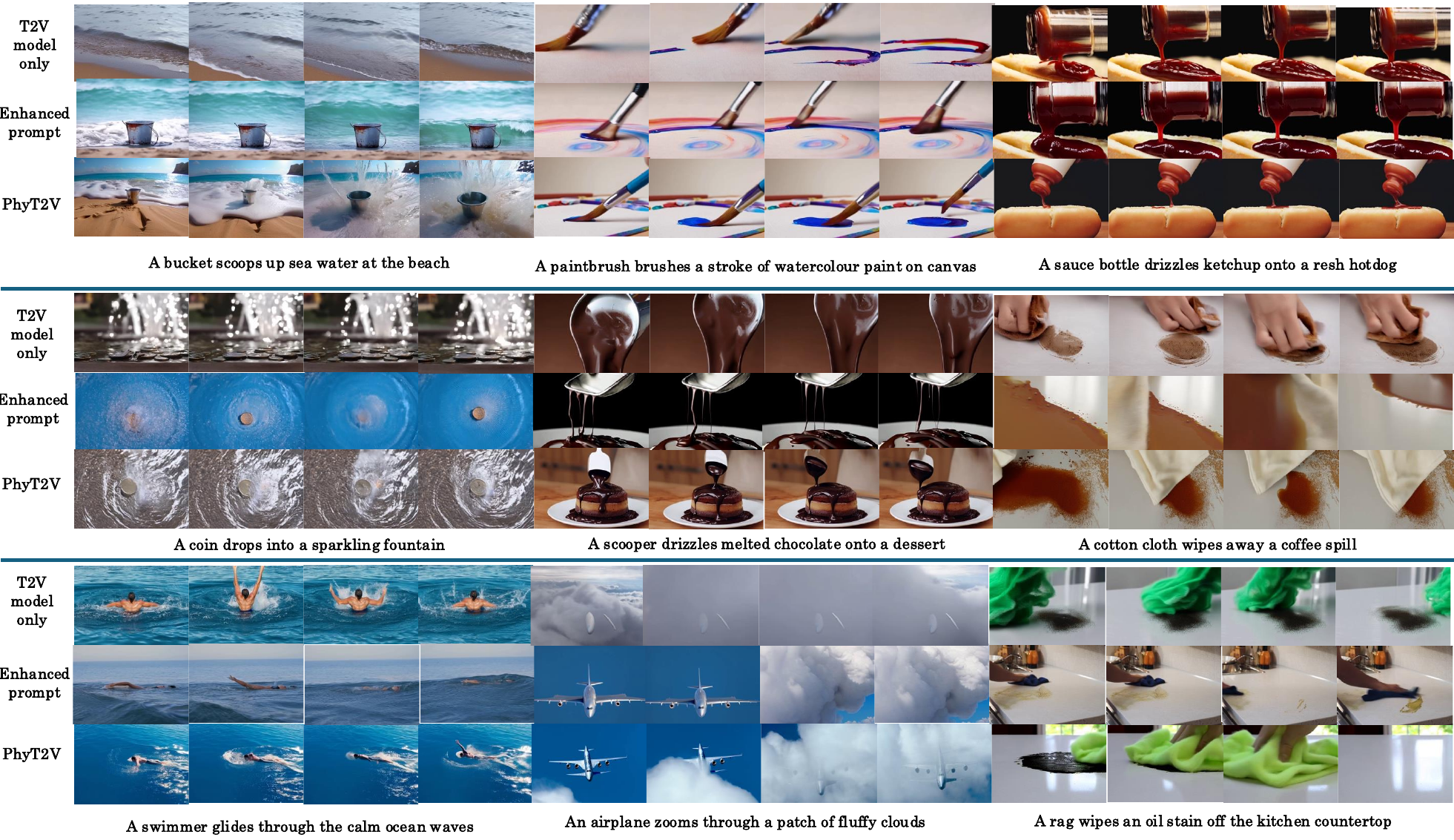} 
\end{figure*}

\begin{figure*}[ht]
	\centering
	\vspace{-1in}  
	\includegraphics[width=\linewidth]{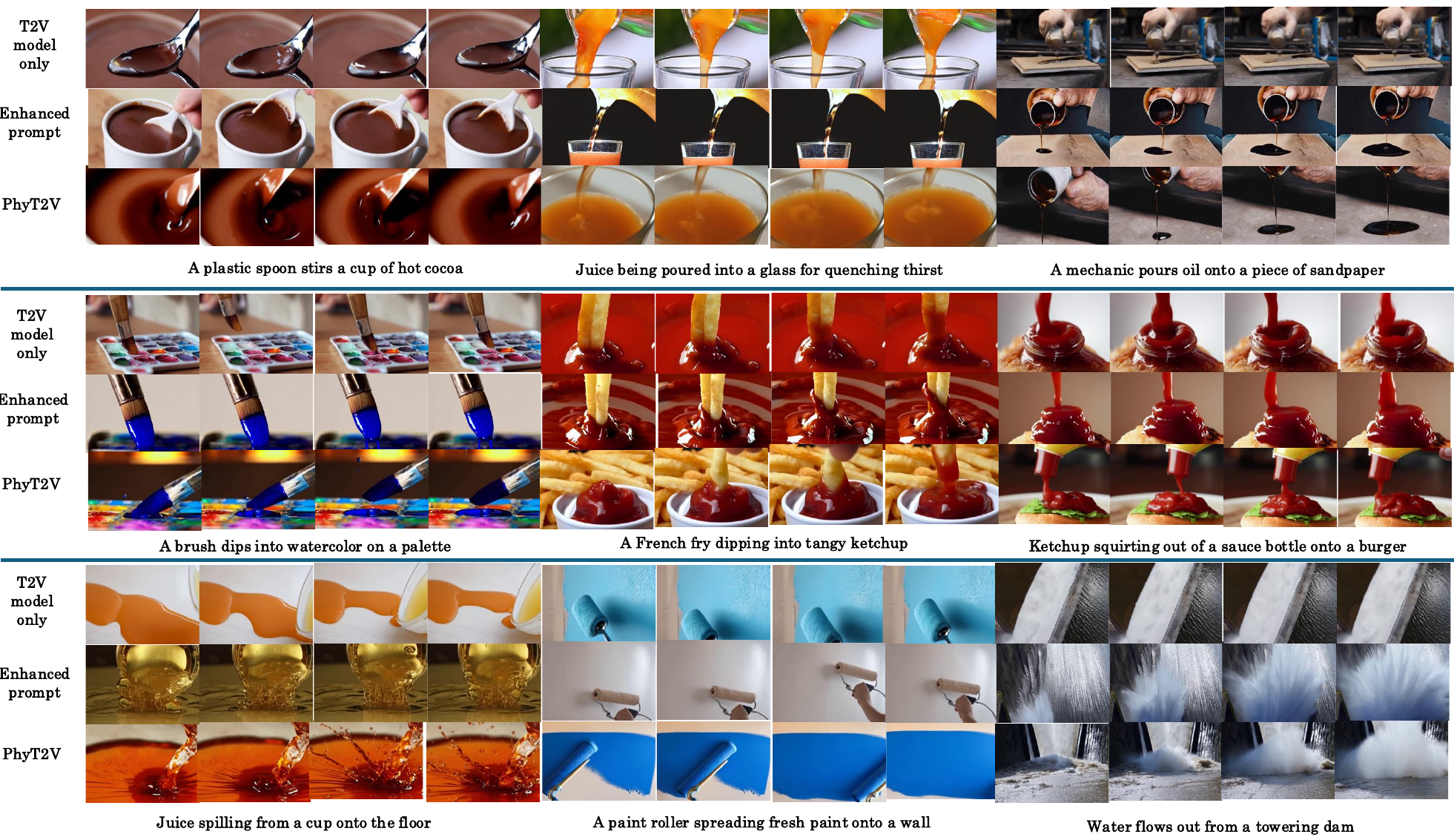} 
\end{figure*}

\begin{figure*}[ht]
	\centering
	\includegraphics[width=\linewidth]{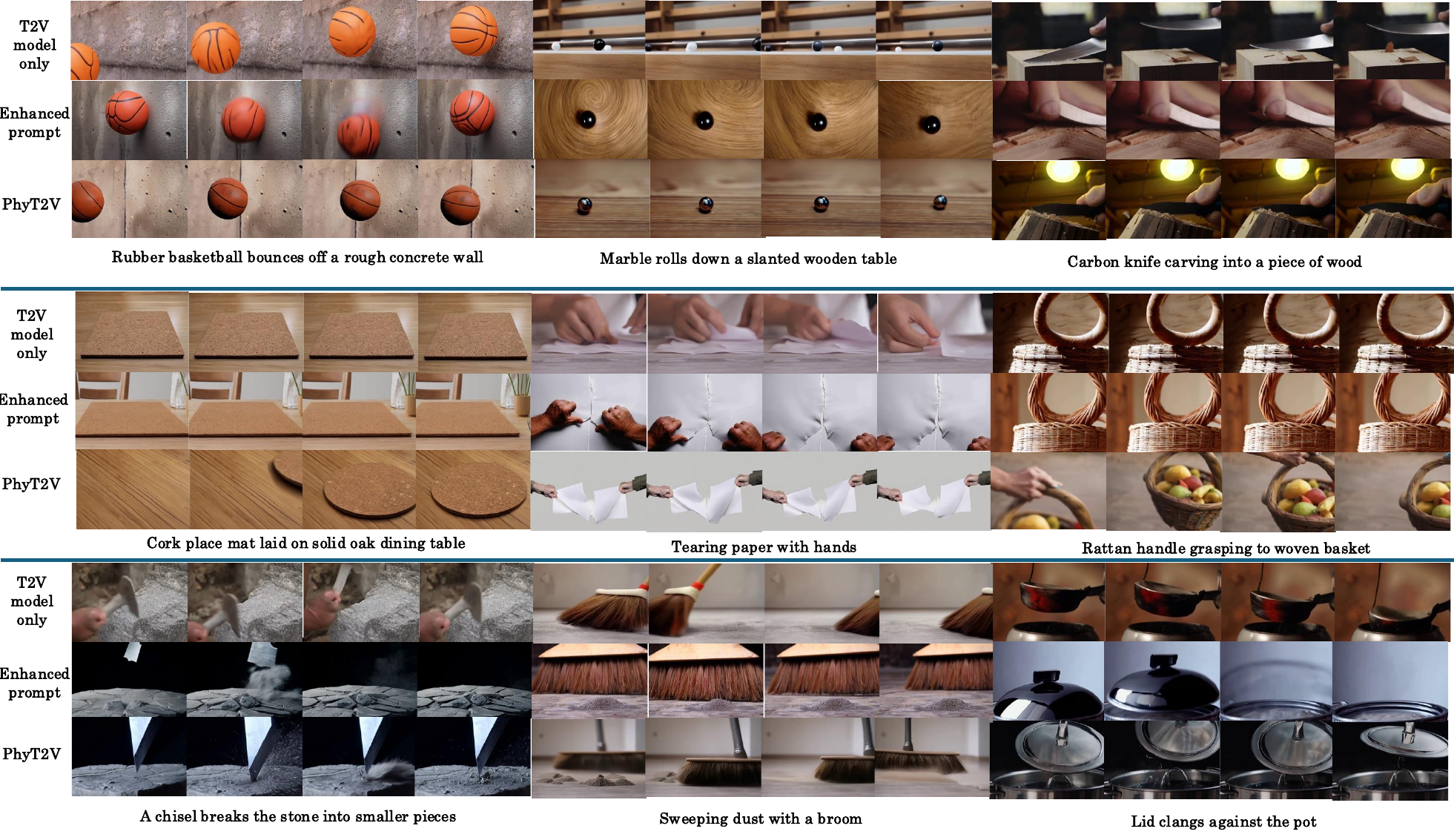} 
\end{figure*}

\begin{figure*}[ht]
	\centering
	\vspace{-1in} 
	\includegraphics[width=\linewidth]{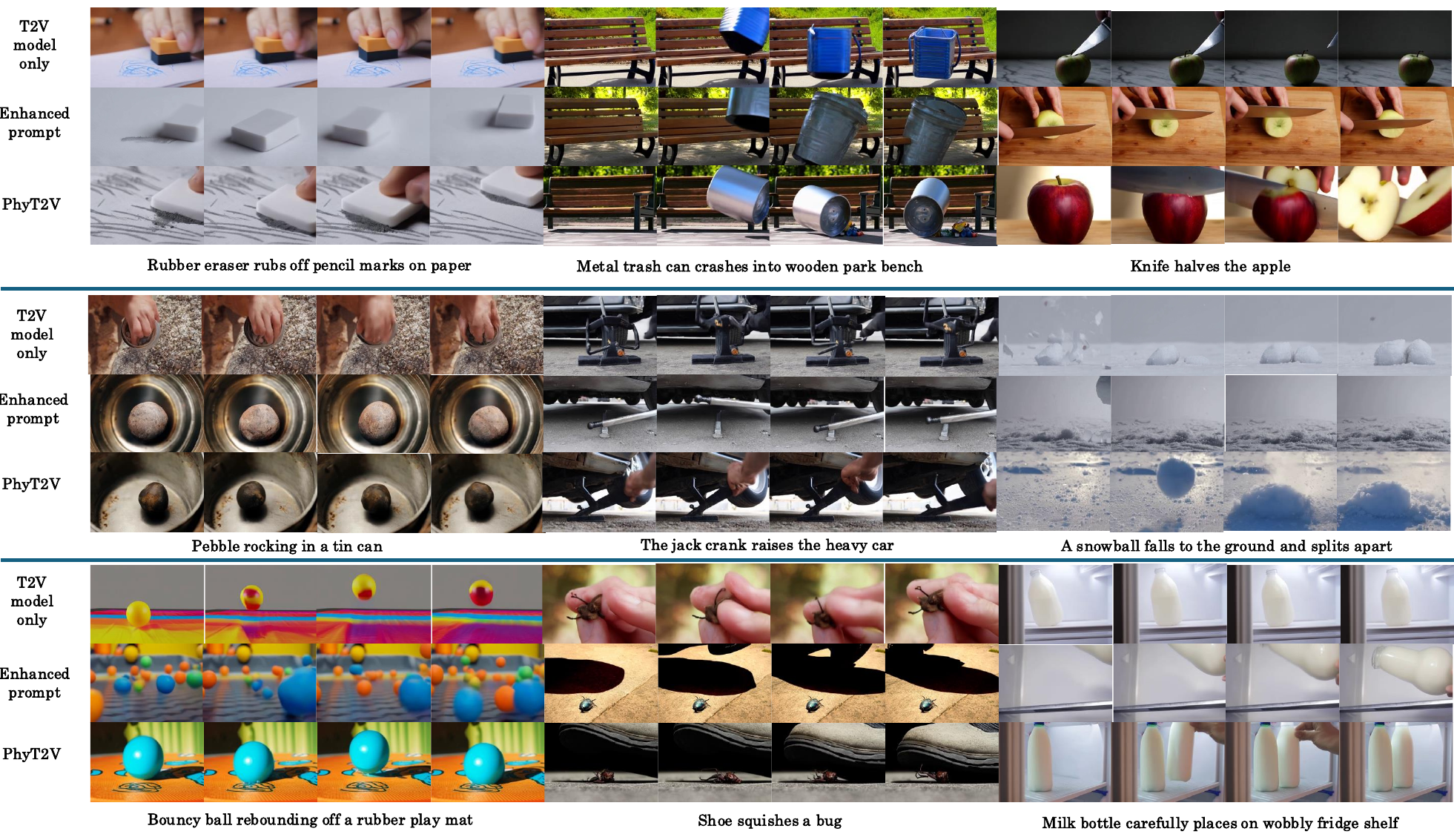} 
\end{figure*}



\end{document}